\documentclass{article}

\usepackage[preprint, eandd]{neurips_2026}

\usepackage[utf8]{inputenc}
\usepackage[T1]{fontenc}

\usepackage{hyperref}
\makeatletter
\newcommand{\thmanchor}[2]{\phantomsection\def\@currentlabel{#1}\label{#2}}
\makeatother
\newcommand{\proofend}{{\unskip\nobreak\hfill\penalty50\hskip1em\null\nobreak\hfill$\square$\parfillskip=0pt\finalhyphendemerits=0\par}}
\hypersetup{colorlinks=true, linkcolor=red!55!black, citecolor=green!45!black, urlcolor=blue!60!black,
  pdftitle={Cross-Dialect Generalization Without Retraining: Benchmarks and Evaluation of Schema-Derived Constrained Decoding for MLIR},
  pdfauthor={Plawan Kumar Rath}}
\usepackage{url}
\usepackage{booktabs}
\usepackage{tabularx}  
\newcolumntype{L}{>{\raggedright\arraybackslash}X}
\usepackage{amsmath}
\usepackage{amssymb}
\usepackage{amsfonts}
\usepackage{nicefrac}
\usepackage{microtype}
\usepackage{xcolor}
\usepackage{graphicx}
\usepackage{listings}
\usepackage{tikz}
\usetikzlibrary{arrows.meta,positioning}
\title{Cross-Dialect Generalization Without Retraining: Benchmarks and Evaluation of Schema-Derived Constrained Decoding for MLIR}

\author{%
  Plawan Kumar Rath\thanks{The views expressed in this paper are those of the author and do not necessarily reflect the views of Meta. This work was conducted in the author's personal capacity.} \\
  Meta \\
  \texttt{plawan@meta.com} \\
}

\begin{document}

\maketitle

\begin{abstract}
Multi-Level Intermediate Representation (MLIR) underlies modern ML compiler infrastructure, including TensorFlow, JAX via StableHLO, PyTorch's Inductor backend, and IREE. Yet it appears only in trace amounts in code-LM pretraining corpora. MLIR is also extensible by design: new dialects ship per application domain, so maintaining a fine-tuned model per dialect does not scale. We ask whether inference-time priors derived mechanically from each dialect's Operation Definition Specification (ODS) can substitute for gradient-based adaptation. We make two contributions. First, we release four natural-language-to-MLIR benchmarks across three dialects (MLIR-Spec-150, Linalg-Spec-30, StableHLO-Spec-30, and StableHLO-Held-Out-200), totaling 410 in-scope NL$\to$MLIR pairs, plus a 25-program StableHLO-OutOfGrammar stress set (435 benchmark instances). Thirty of the Spec prompts, ten per dialect, also carry hand-authored input and expected-output references for functional testing (MLIR-Functional-Reference-30). All artifacts ship under Apache-2.0 with Gebru datasheets and Croissant 1.1 metadata. Second, on top of these benchmarks we build a three-layer schema-derived constraint stack: a context-free grammar over op signatures (C1), type-domain splits from an ODS-extracted type lattice (C2), and an SSA-scope validator driving five-retry rejection sampling (C3). Porting the stack from arith+func+memref+linalg to StableHLO required no new constraint-layer code. Empirically, on dialects whose verifier semantics are dominated by structural constraints, SmolLM2-1.7B with token-level schema-derived constraints matches or exceeds 15B--34B code LMs that run the strongest constraint enforcement their serving stack admits (post-hoc rejection), at a matched retry budget and 8--25$\times$ the per-generation speed: on linalg it reaches 80.0\% verify-valid (three runs, each 80.0\%, $n{=}125$), 21 to 44 pp above CodeLlama-34B, Granite-Code-34B, and StarCoder2-15B with non-overlapping CIs, and the lead survives a same-family fp16 precision control. The comparison is between systems, not models: under the baselines' post-hoc protocol, SmolLM2 itself falls from 52.0\% to 24.5\% on arith+func, a +27.5 pp masking contribution. On arith+func and on the templated parametric StableHLO-Held-Out-200, where verifier semantics turn on attribute values rather than structure, the same baselines match or beat the SLM; we scope these explicitly as non-win cells. We release benchmarks, decoder, every per-prompt generation, and a pinned reproducibility Dockerfile.
\end{abstract}

\section{Introduction}
\label{sec:intro}

MLIR (Multi-Level Intermediate Representation)~\cite{mlir2020} has
become the compiler infrastructure of choice for modern
machine-learning systems, underpinning TensorFlow's compiler stack,
JAX via StableHLO, PyTorch's Inductor backend, and open-source
runtimes such as IREE. Yet MLIR programs appear only in trace amounts
in the public code corpora used to pretrain open-weight LMs, and a
model asked to emit MLIR from natural language routinely produces
grammatically malformed outputs, references operations that do not
exist in the requested dialect, or constructs types that the verifier
rejects on sight. Retraining addresses this in principle, but MLIR is
extensible: the community introduces new dialects per application
domain, and maintaining a fine-tuned model per dialect does not scale.

This paper treats the generation problem as a structural-priors
problem rather than a data problem. MLIR's Operation Definition
Specification (ODS), its TableGen records that declare each
operation's operands, results, type constraints, and structural
invariants, offers a uniform, machine-readable schema shared across
dialects. We ask: on dialects whose verifier semantics are dominated
by structural constraints (operand-rank parity, type-domain splits,
SSA scope), do schema-derived priors let a small LM match or exceed
15B--34B code LMs under matched sampling budgets, and does the
construction port mechanically across dialects? The answer separates:
the derivation transfers across all three dialects without new
constraint-layer code, but the empirical SLM-vs-baseline gap is
dialect- and corpus-conditional and robust on linalg
(structural-dominant), parity-to-loss on arith+func and on templated
parametric StableHLO where attribute-value constraints dominate.

SmolLM2-1.7B-Instruct~\cite{allal2025smollm2} under the full
$C_1+C_2+C_3$ stack, evaluated against
CodeLlama-34B~\cite{roziere2023codellama},
Granite-Code-34B~\cite{mishra2024granite}, and
StarCoder2-15B~\cite{lozhkov2024starcoder2}, each running the
strongest constraint enforcement its serving stack admits under the
same five-retry budget, robustly wins on linalg (80.0\% verify-valid,
mean of three runs at $n{=}125$, each run 80.0\%, $+21$ to $+44$ pp
over baselines with non-overlapping CIs); wins over the 34B baselines on
hand-curated StableHLO-Spec-30;
is overtaken on the templated parametric StableHLO-Held-Out-200 where
the baselines saturate at 98--100\%; and is a non-win cell on
arith+func. Inference on a single Apple M4 Max laptop runs
8--25$\times$ faster per generation than the 34B baselines, with no
fine-tuning, reinforcement learning, or distillation: the entire
adaptation is at inference time. Verify-valid is a structural measure:
under differential execution against the gold module of every spec
prompt, 36.4\% of verify-valid linalg generations compute the correct
function under the canonical interface, and closing that gap is the main
open problem (\S\ref{sec:results}).

We frame this as a datasets-and-evaluation contribution with a
methods contribution that exercises the benchmarks.
(i) \emph{Datasets:} four NL$\to$MLIR benchmarks: MLIR-Spec-150,
Linalg-Spec-30, StableHLO-Spec-30, and StableHLO-Held-Out-200 (the
last a 200-program parametric op-signature sweep that checks against
author-curation bias in Spec-30; 410 in-scope pairs across three
dialects), plus a 25-program StableHLO-OutOfGrammar stress set (435
benchmark instances), all verifier-clean at release
with difficulty tags,
Datasheet-for-Datasets~\cite{gebru2021datasheets} records, and
Croissant 1.1~\cite{croissant2024} metadata under Apache-2.0; to our
knowledge the first published NL$\to$MLIR benchmarks. Thirty Spec
prompts (ten per dialect) also carry hand-authored input and
expected-output references (MLIR-Functional-Reference-30) that lower
each generation to executable IR for output-match testing.
(ii) \emph{Methods:} a schema-derived three-layer constraint pipeline
(CFG over op signatures, type-domain splits from an ODS lattice,
SSA-scope validator with five-retry rejection) extracted mechanically
from ODS; porting from arith+func+memref+linalg to StableHLO, a
dialect with different syntax, different scope semantics, and a
different downstream verifier (\texttt{iree-compile} rather than
\texttt{mlir-opt}), required no new constraint-layer code. The $C_3$
validator admits both a post-hoc rejection-sampling form (used in our
measurements) and an in-line coupled-decoder form, with appendix
proofs of accepted-string-set equivalence.
(iii) \emph{Protocol:} a matched five-retry rejection-sampling budget
applied identically to all models, with the remaining enforcement
asymmetry (token-level masking for the SLM, post-hoc rejection for the
baselines) measured directly rather than assumed away.
(iv) \emph{Findings:} a robust linalg win, a corpus-conditional
StableHLO result, and an honest non-win on arith+func.
Figure~\ref{fig:roadmap} maps benchmarks, constraint layers, and claims
to the evidence that supports each.

\begin{figure}[t]
\centering
\begin{tikzpicture}[font=\footnotesize,
  box/.style={draw, rounded corners=2pt, align=left, inner sep=3pt},
  ar/.style={-{Stealth[length=4pt]}, thick}]
\node[box, text width=4.15cm] (B) {\textbf{Benchmarks} (\S\ref{sec:exp})\\
  MLIR-Spec-150 (arith+func)\\ Linalg-Spec-30 (linalg)\\
  StableHLO-Spec-30\\ StableHLO-Held-Out-200\\
  StableHLO-OutOfGrammar-25\\ MLIR-Functional-Reference-30};
\node[box, text width=3.6cm, right=0.3cm of B] (C) {\textbf{Constraint stack} (\S\ref{sec:method})\\
  $C_1$: grammar of op signatures\\ $C_2$: type-domain splits\\
  $C_3$: SSA-scope rejection\\ SLM: $C_1{+}C_2$ masked\\
  baselines: $C_1{+}C_3$ post hoc};
\node[box, text width=4.3cm, right=0.3cm of C] (R) {\textbf{Claims and evidence}\\
  linalg win (Table~\ref{tab:apples})\\
  layer attribution (Fig.~\ref{fig:progression})\\
  masking effect +27.5 pp (\S\ref{sec:results})\\
  StableHLO: Spec-30 only (Table~\ref{tab:stablehlo_cross_system})\\
  arith+func: non-win (Table~\ref{tab:apples})\\
  functional gap (Table~\ref{tab:verify_vs_functional})};
\draw[ar] (B) -- (C);
\draw[ar] (C) -- (R);
\end{tikzpicture}
\caption{Roadmap. The released benchmarks (left) are generated against a
schema-derived constraint stack (middle); each claim (right) points to
the table or figure that supports it.}
\label{fig:roadmap}
\end{figure}

\section{Related Work}
\label{sec:related}

\paragraph{CFG-guided decoding.} Context-free-grammar-guided decoding
with token-level masks is a mature sub-area of code-model inference.
Willard and Louf~\cite{willard2023outlines} formalize efficient mask
construction via finite-state transitions over regular expressions
and context-free grammars; the Outlines library implements this
approach, and we use it with the llguidance backend. Grammar-Aligned
Decoding~\cite{park2024gad} shows that naive token-level masking
drifts from the CFG-conditioned distribution it is meant to sample
from, and proposes a principled re-normalization. Schall and de
Melo~\cite{schall2025hcs} document a Hidden Cost of Structure: across
several small models, constrained decoding produces lower pass-rates
than free decoding, and the effect is sensitive to prompt format. Our
C1 layer is a standard CFG mask over MLIR operation signatures; what
distinguishes it is that the grammar is generated mechanically from
ODS rather than hand-written, so the same derivation procedure
applies unchanged across dialects. We replicate the
Schall-and-de-Melo setup under three-shot priming with a
bounded-identifier grammar and do not observe the pass-rate reversal
(\S\ref{sec:ablations}); we interpret our ablations as a corroborating
data-point on the prompt-format-sensitivity claim rather than a
rebuttal.

\paragraph{Context-sensitive constrained decoding.} Closer to our C3
layer is a smaller body of work on context-sensitive constrained
decoding for formal targets. Synchromesh~\cite{poesia2022synchromesh}
introduces Constrained Semantic Decoding (CSD), a per-step mask
procedure that uses target-language static analyzers (type checkers,
scope analyzers) to restrict token vocabularies during sampling,
evaluated on SQL and Python synthesis. Type-Constrained Code
Generation~\cite{mundler2025typeconstrained} integrates a static
type-checker into per-step decoding for statically-typed languages
with substantial reductions in type-error rate on a 1B-scale model.
Correctness-Guaranteed Code Generation~\cite{li2025correctness}
extends this direction with a context-sensitive Tree of Parsers that
drives token-level decoding so generated programs are guaranteed
parseable. Our work differs in three respects.
(i) The target is an extensible intermediate-representation family
rather than a general-purpose programming language, and our
C1/C2/C3 layers are derived mechanically from a machine-readable
schema (ODS) rather than hand-implemented against each language's
type system, so a single derivation procedure covers every dialect
that ships ODS records (MLIR core, StableHLO, and any out-of-tree
dialect that follows the same convention).
(ii) We prove equivalence between the in-line coupled decoder and
post-hoc rejection sampling on the same accepted-string set, with a
BPE-boundary argument that handles subword tokens spanning multiple
grammar terminals (Theorem~\ref{thm:eq}, Appendix~\ref{app:soundness}).
(iii) We release four NL$\to$MLIR benchmarks; Synchromesh and the
type-constrained line evaluate on existing code benchmarks but do
not release target-specific datasets.

\paragraph{Small language models for formal targets.} Language-model
approaches to formal targets have concentrated on proof assistants as
shown by Lean tactic generation~\cite{yang2023leandojo} and Isabelle
whole-proof synthesis~\cite{first2023baldur}, and on
hardware-description languages such as Verilog~\cite{liu2023verilogeval}.
MLIR has received comparatively little attention as a direct generation
target despite its central role in compiler and ML-systems
infrastructure; to our knowledge this is the first work to treat MLIR
as a training-free language-model generation target at the
dialect-agnostic level. Our contribution is training-free; fine-tuning
is an orthogonal axis we do not exercise here but which would compose
with the schema-derived stack.

\section{Method}
\label{sec:method}

\paragraph{Pipeline overview.}
Our method takes a natural-language description $x$ and produces MLIR
code $y$ by sampling from a small language model under a three-layer
constraint stack $C = (C_1, C_2, C_3)$ derived mechanically from the
target dialect's Operation Definition Specification (ODS). The three
layers cover distinct aspects of MLIR well-formedness: $C_1$ is a
context-free grammar over operation signatures, $C_2$ splits that
grammar by operand and result type domain, and $C_3$ is a dynamic
validator that enforces SSA-scope correctness. $C_1$ and $C_2$ are
applied as token-level masks during decoding; $C_3$ drives a five-retry
rejection sampler around the masked decoder. The same derivation
procedure applies unchanged across dialects: porting the stack from
arith+func+memref+linalg to StableHLO required no new
constraint-layer code.

\paragraph{Schema extraction.}
We extract ODS records via \texttt{llvm-tblgen --dump-json} over the
LLVM, IREE, and StableHLO TableGen source trees, yielding 276 operation
records across the dialects we target. Each record names an operation's
operands, results, attributes, and structural traits (e.g.,
\texttt{SameOperandsAndResultType}, \texttt{ElementwiseMappable}). We
flatten the multi-inheritance class hierarchies and produce two derived
structures: (i) a per-dialect operation lattice naming operand and
result type constraints, and (ii) a per-dialect attribute lattice
naming attribute kinds and value domains. Both are consumed by the
$C_1$ grammar generator and the $C_2$ splitter. (TableGen's JSON dump
resolves the multi-inheritance hierarchy correctly; an earlier
regex-based parser resolved only 2 of the 98 linalg operations we
need.)

\paragraph{$C_1$: context-free grammar over op signatures.}
$C_1$ is a LALR-compiled context-free grammar over the MLIR surface
syntax restricted to the target dialect's operation set. Productions
are generated from the flattened ODS: each operation contributes a
production naming its mnemonic, operand arity, attribute slots, and
result type constraints in the concrete MLIR syntax (e.g.,
\texttt{arith.addi} \mbox{\texttt{\%a, \%b :\ i32}} rather than the ODS declaration
form). The grammar is compiled to a token-level automaton and delivered
as a logits-level mask via Outlines~\cite{willard2023outlines} with its
llguidance backend. Two engineering choices are critical: explicit
whitespace productions (replacing LARK's default \texttt{\%ignore WS},
which silently exhausts \texttt{max\_tokens} without grammar progress)
and bounded SSA identifier regexes (preventing runaway names). With
these $C_1$ produces 96--99\% parse-valid output across dialects.

\paragraph{$C_2$: type-domain grammar splits.}
$C_2$ refines $C_1$ by splitting productions along type domains read
from the ODS type lattice. For example, \texttt{arith.addi} is
constrained to integer operand and result types, \texttt{arith.addf}
to floating-point types, and \texttt{arith.cmpi} takes the integer
comparison predicate vocabulary (\texttt{eq, ne, slt, \ldots}) while
\texttt{arith.cmpf} takes the floating-point predicate vocabulary
(\texttt{oeq, olt, une, \ldots}). In linalg the splits encode richer
constraints: \texttt{linalg.matmul} on
\texttt{memref<M$\times$K$\times$f32>} by
\texttt{memref<K$\times$N$\times$f32>} requires the inner dimension to
be consistent across operands; \texttt{linalg.transpose} constrains
the permutation attribute to match the source memref rank. Each split
is determined mechanically from the ODS record's trait set (e.g.,
\texttt{SameOperandsAndResultType} implies a single type domain across
all positions), and we do not hand-author splits per dialect.

\paragraph{$C_3$: SSA-scope validator.}
$C_3$ is a dynamic validator over SSA uses. Given a generated MLIR
fragment, it parses line-by-line, maintains a per-function symbol
table seeded with the function's parameter list, updates the table at
each SSA definition site, and validates each use site against the
conjunction (name-in-scope $\land$ type-matches-declaration). On a
validation failure the fragment is rejected and the decoder retries.
We use a five-retry budget: the first attempt samples greedily
(temperature 0) so it coincides with a pure $C_1 + C_2$ decode on
accept; subsequent retries use temperature 0.8 to sample a different
completion. On $n{=}200$ SmolLM2 $C_1$ arith+func generations from the
frozen released pool, the validator's confusion matrix against
\texttt{mlir-opt --verify-diagnostics} shows 3 false rejects among 89
verify-valid programs (Appendix Fig.~\ref{fig:scope}); on the $C_1$ outputs of
every 15B/34B baseline the count is zero. All SmolLM2 false rejects come
from operations that the grammar allows to span several lines, which the
line-based scope tracker misreads (Appendix Table~\ref{tab:c3_false_rejects}).

\paragraph{Formal guarantees.}
Appendix~\ref{app:soundness} proves three theorems about an in-line
reformulation of $C_3$ as a coupled state machine: soundness
(Theorem~\ref{thm:sound}), BPE-refined coverage
(Theorem~\ref{thm:cov}), and equivalence to post-hoc rejection sampling
(Theorem~\ref{thm:eq}). Together they guarantee that the token-masked
decoder never emits a string outside the grammar's support, and that the
in-line and post-hoc forms of $C_3$ accept exactly the same programs.
Their practical value is attribution: any difference in \emph{what} gets
accepted comes from the constraint set, not from one decoder's
implementation. Masking still changes success rates by reshaping the
proposal distribution, which \S\ref{sec:results} measures. The theorems
do not guarantee task success, verifier acceptance, or agreement between
$C_3$ and \texttt{mlir-opt}; that agreement is measured empirically in
Appendix~\ref{app:soundness}. All empirical results use the post-hoc
form; \S\ref{sec:limits} and \S\ref{sec:future} report the in-line
implementation's empirical gap, and Appendix~\ref{app:notation} collects
the notation.

\section{Experimental Setup}
\label{sec:exp}

\subsection{Models}
Primary: SmolLM2-1.7B-Instruct~\cite{allal2025smollm2} (MLX fp16).
Baselines: CodeLlama-34B~\cite{roziere2023codellama} Q4\_K\_M,
Granite-Code-34B~\cite{mishra2024granite} Q4\_K\_M, and
StarCoder2-15B-Instruct~\cite{lozhkov2024starcoder2} Q4\_0
(all via Ollama / llama.cpp). The Outlines/llguidance backend used for
C1+C2~\cite{willard2023outlines} is fixed across all SmolLM2 cells.

\subsection{Benchmarks}
Evaluating a natural-language-to-MLIR generation system requires paired
(NL, MLIR) data that, to our knowledge, is not available in any published
benchmark. Recent NL-to-code benchmarks released through different venues
such as HumanEval~\cite{chen2021humaneval}, MBPP~\cite{austin2021mbpp},
CodeContests~\cite{li2022alphacode}, APPS~\cite{hendrycks2021apps},
MultiPL-E~\cite{cassano2023multipl}, pair natural-language problem
statements with programs in general-purpose programming languages, and
none pairs NL with MLIR or any other production compiler intermediate
representation. We therefore construct four benchmarks from scratch,
spanning three dialects, plus a 25-program out-of-grammar stress set;
Table~\ref{tab:benchmarks} summarizes all five.\footnote{Code: \url{https://github.com/plawanrath/slm-to-mlir-constrained-emitter}. Datasets (Hugging Face collection): \url{https://huggingface.co/collections/plawanrath/cross-dialect-mlir-benchmarks-neurips-2026-e-and-d}.}

\textbf{MLIR-Spec-150}: 150 hand-authored NL$\to$MLIR pairs,
arith+func+memref, difficulty mix 38/44/18\% easy/medium/hard, all
verify-clean. \textbf{Linalg-Spec-30}: 30 hand-authored
pairs for the 12 linalg named ops in scope
(matmul, matvec, fill, copy, transpose, broadcast, add, sub, mul, div, exp,
abs) under memref semantics. \textbf{StableHLO-Spec-30}: 30 hand-authored
StableHLO~\cite{stablehlo2024} pairs covering 10 op families.
\textbf{StableHLO-Held-Out-200}: 200 programs generated by a
parametric sweep over the StableHLO opset (7 op families $\times$
6 dtypes $\times$ multiple shape ranks; 585 raw candidates filtered
to 200 \texttt{iree-compile}-clean~\cite{iree2024}; see
\S\ref{sec:results}) which is a held-out corpus that reduces
author-design bias in the StableHLO evaluation.
\textbf{StableHLO-OutOfGrammar-25}: 25
programs using StableHLO ops outside our 10-op grammar, used to
characterize graceful-degradation behavior. Thirty Spec prompts, ten
from each of MLIR-Spec-150, Linalg-Spec-30, and StableHLO-Spec-30, also
carry hand-authored input and expected-output references
(MLIR-Functional-Reference-30, \texttt{eval/functional/}). These
references add no new prompts; they annotate existing instances for
output-match testing and are released as evaluation evidence
(\S\ref{sec:results}).

\textbf{Evaluation pools.} The arith+func and linalg results use two
fixed pools, frozen at release and identical across every cell and run.
The arith+func pool ($n{=}200$) is the 150 MLIR-Spec prompts plus 50
padding prompts drawn by a fixed seed-0 shuffle from weak
natural-language descriptions of LLVM/MLIR test files. The linalg pool
($n{=}125$) is the 30 Linalg-Spec prompts plus all 95 corpus prompts
that pass the eligibility filters (the 12 target ops, memref semantics,
non-empty description), so 125 is corpus exhaustion. Spec and padding
prompts are disjoint; the mined descriptions are templated, so the pools
hold 193 and 75 unique strings. Tables report the all-prompt rate;
Table~\ref{tab:unique} repeats every pooled cell with one value per
unique prompt, and the released per-prompt files carry provenance labels
(spec or padding, duplicate group). Conclusions are unchanged under
either aggregation.

\begin{table}[h]
\centering
\caption{Benchmark summary: the four released benchmarks plus the
out-of-grammar stress set. Total: 410 in-scope NL$\to$MLIR pairs
across three dialects, with a 25-program out-of-scope supplement.}
\label{tab:benchmarks}
\small
\setlength{\tabcolsep}{3.5pt}
\begin{tabularx}{\linewidth}{@{}lr>{\raggedright\arraybackslash}p{1.55cm}Lllc@{}}
\toprule
Benchmark                   & $n$ & Dialect          & Ops covered                  & E/M/H        & Source & Verifier      \\
\midrule
MLIR-Spec-150               & 150 & arith+\allowbreak func+\allowbreak memref & ${\sim}50$ distinct          & 38/44/18\%   & hand   & mlir-opt      \\
Linalg-Spec-30              &  30 & linalg (memref)   & 12 named                     & -          & hand   & mlir-opt      \\
StableHLO-Spec-30           &  30 & stablehlo         & 10 op families               & 40/40/20\%   & hand   & iree-compile  \\
StableHLO-Held-Out-200      & 200 & stablehlo         & 7 fam$\times$6 dtypes$\times$ranks & sweep  & sweep  & iree-compile  \\
\midrule
StableHLO-OutOfGrammar-25 &  25 & stablehlo         & 23 fam, out of grammar       & -          & hand   & iree-compile  \\
\bottomrule
\end{tabularx}
\end{table}

Reuse scenarios beyond this paper's protocol are itemized in the
datasheet (Appendix~\ref{sec:datasheet}).

\subsection{Statistics}
Paired bootstrap (10\,000 resamples, 95\% CI percentile method); the
resampling unit is the prompt row, paired by (run, prompt\_id). All
tables report $n$ per cell.

\paragraph{Power analysis.} For unpaired cross-system comparisons at
$n=200$ (arith+func, baseline rate $p=0.55$) the minimum detectable
effect (MDE) at $\alpha=0.05$, $1-\beta=0.80$ (two-proportion normal
approximation) is ${\sim}9.8$pp, and at $n=125$ (linalg, $p=0.68$)
${\sim}11.4$pp; cross-system differences below these bounds are reported
as null. Constraint-layer deltas within one system are paired over
identical prompts and are judged by their paired-bootstrap CIs, which
have more power: the arith+func $C_3$ increment ($+6.5$pp, CI $[+2.0,
+11.0]$) excludes zero, while the linalg $C_3$ increment ($+4.0$pp, CI
$[+0.0, +8.0]$) reaches zero and is not reported as significant.

\subsection{30B baseline protocol}
\texttt{C1} on Ollama is implemented as rejection sampling against the
LARK parse grammar with up to 5 retries, because llama.cpp exposes no
logit-mask hook for LARK grammars; $C_3$ applies the same scope acceptor.
This is weaker enforcement than SmolLM2's token-level masking, and
\S\ref{sec:results} measures the difference directly. Baselines decode at
temperature 0.2 on the first attempt and 0.3 on retries with
\texttt{max\_tokens=256}; rerunning them at SmolLM2's 600-token budget
leaves every cell statistically unchanged
(Appendix~\ref{sec:reproducibility-appendix}).

\section{Results}
\label{sec:results}

\paragraph{Headline metrics.} On the frozen released pools, the seed-0
constraint ladder for SmolLM2-1.7B reads $20.0 \to 44.5 \to 45.5 \to
52.0\%$ on arith+func ($n{=}200$) and $52.0 \to 64.0 \to 73.6 \to
77.6\%$ on linalg ($n{=}125$) for free $\to C_1 \to C_1{+}C_2 \to
C_1{+}C_2{+}C_3$ (Appendix Table~\ref{tab:main}). The linalg top rung lies within
the measured run-to-run variation (at most 2.4pp) of the three-run
80.0\% below: prompts resolved by the greedy first attempt reproduce
byte-identically across runs, and all variation comes from the
temperature-0.8 retries. Without $C_3$, five-retry $C_1$ rejection
sampling gives CodeLlama-34B 60.5\% / 56.8\% and Granite-Code-34B
47.0\% / 33.6\% on the same pools.

\paragraph{Apples-to-apples comparison.} Table~\ref{tab:apples}
reports the matched-budget five-retry $C_1+C_3$ comparison at
$n{=}200$ arith+func and $n{=}125$ linalg, three runs with paired
prompts; a Granite-Code-8B-fp16 cell controls for precision. On linalg,
SmolLM2 attains 80.0\% (each of three runs at 80.0\%), exceeding
CodeLlama-34B (58.7\%) by
$+21.3$pp, Granite-34B (35.7\%) by $+44.3$pp, and StarCoder2-15B
(54.9\%) by $+25.1$pp with non-overlapping CIs across all four
systems; the fp16 8B control trails by 29.9pp, so the gap is not a
quantization artifact. On arith+func, SmolLM2
(53.2\%, $\pm 1.8$pp) is within CI of CodeLlama-34B (59.8\%) and
Granite-34B (51.5\%), trails StarCoder2-15B (66.8\%) by 13.6pp with
non-overlapping CIs, and is dominated by Granite-Code-8B-fp16
(68.2\%, paired $\Delta$ $+15.0$pp, $p<0.001$) and we scope arith+func
explicitly as a non-win cell. The shared $C_3$ filter does not
penalize any baseline: without it, the baselines score 60.5\%
(CodeLlama-34B), 67.0\% (StarCoder2-15B), and 47.0\% (Granite-34B) on
arith+func, within noise of or below their $C_3$-gated rates, and on
linalg $C_3$ helps every baseline ($+1.9$ to $+7.7$pp). Against each
baseline's best configuration (Table~\ref{tab:bestconfig}), SmolLM2's
linalg lead is at least $+21.3$pp.

\paragraph{Constraint setting and the converse cell.} The comparison is
system-level. SmolLM2 runs the strongest constraint implementation its
stack admits (token-level $C_1{+}C_2$ masking with $C_3$ rejection); the
15B/34B baselines run the strongest theirs admits (post-hoc $C_1{+}C_3$
rejection), at a matched five-retry budget. We measure this enforcement
asymmetry on the one model that supports both protocols: SmolLM2 under
the baselines' post-hoc protocol scores 24.5\% on arith+func (27.0\% at
the baselines' 256-token budget and temperatures) versus 52.0\% with
token-level masking, a paired masking contribution of $+27.5$pp (95\%
CI $[+20.0, +35.0]$). The gains therefore come from schema-derived
enforcement rather than from the small model alone: without masking,
SmolLM2 falls below every 15B--34B baseline on this dialect.

\begin{table}[h]
\centering
\caption{Apples-to-apples comparison under a matched five-retry budget,
mean of three runs ($n{=}200$ per run arith+func; $n{=}125$ per run
linalg); across-run half-range in parentheses. Granite-Code-8B-fp16 is a
same-family precision control on both dialects (\S\ref{sec:limits}).
\emph{Reading guide:} compare each baseline row with SmolLM2 per
dialect; linalg is the win and arith+func the scoped non-win.}
\label{tab:apples}
\small
\begin{tabular}{lrr}
\toprule
System                                            & arith+func           & linalg                       \\
\midrule
SmolLM2-1.7B + $C_1$+$C_2$+$C_3$                  & 53.2\% ($\pm$1.8pp)  & \textbf{80.0\% ($\pm$0.0pp)} \\
CodeLlama-34B + $C_1$+$C_3$                       & 59.8\% ($\pm$0.8pp)  & 58.7\% ($\pm$2.4pp)          \\
Granite-Code-34B + $C_1$+$C_3$                    & 51.5\% ($\pm$2.5pp)  & 35.7\% ($\pm$4.4pp)          \\
StarCoder2-15B + $C_1$+$C_3$                      & 66.8\% ($\pm$1.5pp)  & 54.9\% ($\pm$1.2pp)          \\
\midrule
Granite-Code-8B-fp16 + $C_1$+$C_3$ (fp16 control) & 68.2\% ($\pm$3.0pp)  & 50.1\% ($\pm$1.6pp)          \\
\bottomrule
\end{tabular}

\end{table}

\paragraph{Cross-IR generalization (StableHLO).} Porting the same
pipeline mechanically to StableHLO~\cite{stablehlo2024} (verifier:
\texttt{iree-compile --iree-input-type=stablehlo --compile-to=input}~\cite{iree2024};
toolchain status in Appendix~\ref{app:stablehlo_toolchain}) without
new constraint-layer code, SmolLM2 + $C_1+C_3$ verifies at 63.3\% on
hand-curated Spec-30 ($n{=}30$) and 61.5\% on the parametric
Held-Out-200 ($n{=}200$). The two corpora diverge sharply for the
15B/34B baselines (Table~\ref{tab:stablehlo_cross_system}): on
Spec-30, SmolLM2 leads the 34B baselines by $+26.6$ to $+30.0$pp and
sits within bootstrap CI of StarCoder2-15B; on Held-Out-200 the same
baselines saturate at 98--100\% and SmolLM2 trails by $-36.5$ to
$-38.5$pp with paired-bootstrap CIs excluding zero. We read this as
a benchmark-design finding: hand-curated Spec-30 prompts stress the
baselines while parametric Held-Out-200 templates do not, so Spec-30
is the discriminating corpus and Held-Out-200 is most useful as a
within-model robustness probe; the SLM-vs-baseline lead is retained
on Spec-30 and not extended to Held-Out-200. On OutOfGrammar-25,
$C_1+C_3$ correctly emits 0/25 (the grammar cannot produce ops it
does not model which is the system's intended failure mode), while free
decoding produces 3/25, reflecting pretraining leakage rather than a
property of the constrained system.

\begin{table}[h]
\centering
\caption{StableHLO cross-system comparison. SmolLM2-1.7B at full
stack ($C_1+C_3$) vs the three Ollama baselines under matched
$C_1+C_3$ with 5-retry rejection, on the three released StableHLO
benchmarks. SmolLM2 constraint-ladder on StableHLO:
$26.7 \to 63.3 \to 63.3\%$ (Spec-30) and
$44.0 \to 57.0 \to 61.5\%$ (Held-Out-200) for
free $\to$ $C_1$ $\to$ $C_1+C_3$; the $+C_2$ layer is silent on
StableHLO because operand types are constrained structurally rather
than by named domain split. \emph{Reading guide:} Spec-30
discriminates between systems, while Held-Out-200 saturates the
baselines, so the lead holds on Spec-30 only.}
\label{tab:stablehlo_cross_system}
\small
\setlength{\tabcolsep}{5pt}
\begin{tabular}{lrrrr}
\toprule
Benchmark & \begin{tabular}[b]{@{}r@{}}SmolLM2\\1.7B\end{tabular} & \begin{tabular}[b]{@{}r@{}}CodeLlama\\34B\end{tabular} & \begin{tabular}[b]{@{}r@{}}Granite-Code\\34B\end{tabular} & \begin{tabular}[b]{@{}r@{}}StarCoder2\\15B\end{tabular} \\
\midrule
Spec-30 ($n{=}30$, hand-curated)       & \textbf{63.3\%} & 36.7\%        & 33.3\%           & 60.0\%         \\
Held-Out-200 ($n{=}200$, parametric)   & 61.5\%          & 100.0\%       & 98.0\%           & 100.0\%        \\
OutOfGrammar-25 ($n{=}25$, stress)   & 0\% (intended)  & ---           & ---              & ---            \\
\bottomrule
\end{tabular}

\end{table}

\paragraph{Verify-valid vs functional correctness.} Verify-valid is a
structural metric, and the functional evidence is weaker than the
headline. On the hand-authored functional references (one fixed input
each), output-match is 80\% on arith+func, 20\% on linalg, and 50\% on
StableHLO (Table~\ref{tab:verify_vs_functional}). To test at benchmark
scale without changing the released data, a gold-differential harness
(released as \texttt{eval/functional\_differential/}) executes the frozen
seed-0 generations against the gold module that ships with each of the
180 MLIR-Spec-150 and Linalg-Spec-30 prompts, on five seeded random input
draws, after a two-run gold-vs-gold determinism gate that 173 of 180
golds pass. Of the verify-valid generations, 43/68 = 63.2\% on
arith+func and 8/22 = \textbf{36.4\% on linalg} compute the gold's
function under the canonical interface (43/60, CI $[60.0, 83.3]$, and
8/9 among signature-compatible generations). The distance between
linalg's 80.0\% verify-valid and 36.4\% canonical-interface correctness
is the main limit on reading the headline as capability.

\begin{table}[h]
\centering
\caption{Verify-valid vs functional correctness for SmolLM2 +
$C_1$+$C_2$+$C_3$. Verify-valid is the three-run mean on the evaluation
pool. Fixed-input output-match is on the hand-authored functional
references (10 per dialect). Gold-differential is the share of
verify-valid seed-0 generations on the spec prompts that match the gold
module on all five random draws; non-conforming interfaces count as
failures. The StableHLO golds are not part of the differential run.
\emph{Reading guide:} read across a row to see how much of the
structural rate survives execution.}
\label{tab:verify_vs_functional}
\small
\begin{tabularx}{\linewidth}{@{}Lrrr@{}}
\toprule
Dialect & \begin{tabular}[b]{@{}r@{}}Verify-valid\\(three runs)\end{tabular} & \begin{tabular}[b]{@{}r@{}}Fixed-input output-match\\(10 references)\end{tabular} & \begin{tabular}[b]{@{}r@{}}Gold-differential\\(spec prompts)\end{tabular} \\
\midrule
arith+func          & 53.2\% ($n{=}200$) & 80.0\% (8/10) & 63.2\% (43/68) \\
linalg              & 80.0\% ($n{=}125$) & 20.0\% (2/10) & 36.4\% (8/22)  \\
StableHLO (Spec-30) & 63.3\% ($n{=}30$)  & 50.0\% (5/10) & n/a            \\
\bottomrule
\end{tabularx}

\end{table}

Much of the linalg shortfall is interface non-conformance rather than
wrong computation: 13 of the 22 verify-valid linalg generations declare
a different interface than the gold (dropped output arguments, wrong
rank or element type), which a textual verifier cannot see, and every
non-matching scored generation lowers and executes, so its failure is
semantic. The most direct evidence for this diagnosis is the smaller
linalg reference experiment ($n{=}10$): re-executing its generations
under wrappers adapted to each generation's declared signature (rank or
argument count) moves the cascade from 8/10 verify-valid, 4/10 lowered,
3/10 executed, and 2/10 matching to 8/10, 8/10, 8/10, and 7/10; the one
remaining failure computes a matrix product instead of an elementwise
product. Randomized inputs also sharpen the fixed-input check: across
150 trials on the references, every fixed-input pass holds on all five
draws, 17 of the 20 scorable candidates match on every draw, and one
semantic failure matches on exactly 1 of 5 draws, the input-dependent
case a single fixed input can miscount.

\paragraph{Efficiency frontier.} SmolLM2 $C_1+C_2+C_3$ runs at
1.65--1.86~s/gen on M4 Max vs ${\approx}16$~s (CodeLlama-34B + $C_1$)
and ${\approx}40$~s (Granite-34B + $C_1$) under the same 5-retry
budget which is 8--25$\times$ faster, at higher verify-valid on linalg
and lower on arith+func. Full frontier (Fig.~\ref{fig:efficiency}),
qualitative example, and summary-of-claims
(Table~\ref{tab:empirical_claims}) are in Appendix~\ref{app:extras}.

\section{Ablations}
\label{sec:ablations}

\begin{figure}[h]
\centering
\includegraphics[width=\linewidth]{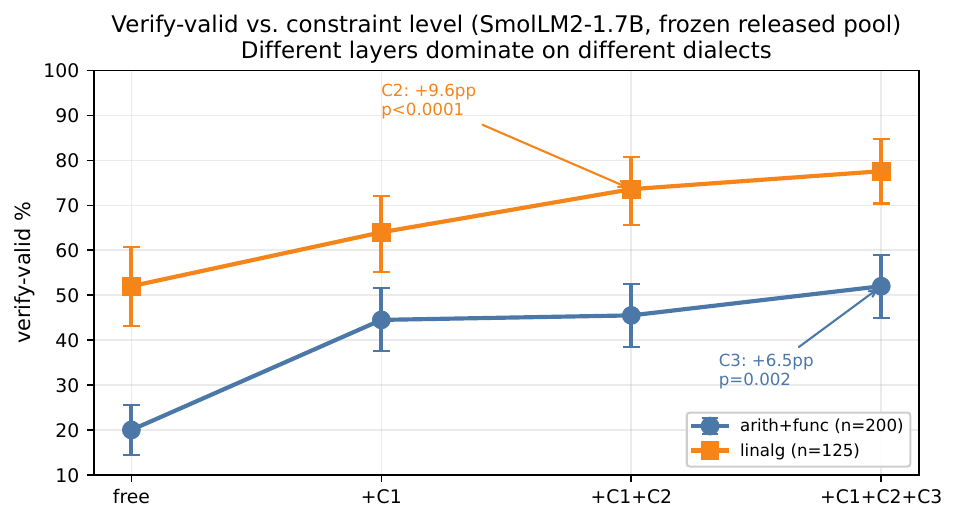}
\caption{SmolLM2-1.7B verify-valid vs constraint level on the frozen
released pools. Different layers dominate on different dialects: $C_3$
drives the arith+func lift ($+6.5$pp, $p=0.002$); $C_2$ drives the
linalg lift ($+9.6$pp, $p<0.0001$). Error bars are 95\% bootstrap CIs.
\emph{Reading guide:} the slope between adjacent points is the paired
effect of adding one layer.}
\label{fig:progression}
\end{figure}

Figure~\ref{fig:progression} shows SmolLM2's verify-valid rate at each
constraint level on both dialects. Supporting figures (error-category progression, paired-bootstrap
deltas, HCS replication) and the per-error-category table are in
Appendix~\ref{app:extras}.

\paragraph{C1$\to$C1+C2 on linalg: $+9.6$pp verify} (paired bootstrap,
CI $[+4.8, +15.2]$, $p<0.0001$). The $C_2$ effect concentrates on linalg,
whose ins/outs structural typing benefits from type-domain splits; on
arith+func $C_2$ adds $+1.0$pp (CI $[+0.0, +2.5]$), a null.

\paragraph{C1+C2$\to$C1+C2+C3 on arith+func: $+6.5$pp verify} (paired
bootstrap, CI $[+2.0, +11.0]$, $p=0.002$). $C_3$ drives the arith+func
lift; on linalg it adds $+4.0$pp (CI $[+0.0, +8.0]$, not significant)
because the target error class (cross-SSA type) is already rare after
$C_2$ (Fig.~\ref{fig:err}).

\paragraph{Hidden Cost of Structure (HCS).} We replicate the RANLP 2025
setup on the frozen arith+func pool. In our stack (3-shot priming +
bounded-identifier grammar + type-domain splits), \textbf{HCS reversal is
not observed}: $C_1$ is monotone beneficial on both SmolLM2 ($+24.5$pp vs
free, CI $[+17.5, +31.5]$) and Phi-3.5-mini ($+20.0$pp, CI $[+14.0,
+26.0]$), both $p<0.0001$ (Fig.~\ref{fig:hcs}). The phenomenon appears
setup-conditional; modern constraint engineering absorbs it.

\section{Limitations}
\label{sec:limits}

\paragraph{30B ``C1'' is rejection sampling, not token-level masked
decoding.} llama.cpp does not accept LARK grammars, so our 15B/34B
baselines (CodeLlama, Granite-Code, StarCoder2 via Ollama) generate
freely and accept the first of 5 retries that parses under the LARK
coverage grammar which is a weaker constraint enforcement than SmolLM2's true
token-masked $C_1$ via Outlines/llguidance. Closing this would
require integrating llguidance into llama.cpp or running a 15B/30B
model through MLX, neither of which is on the single-laptop budget
for this paper. We therefore measure the asymmetry ($+27.5$pp masking
contribution, \S\ref{sec:results}), and every small-versus-large claim
in this paper is a system-level comparison.

\paragraph{Verify-valid $\ne$ functional correctness.} Verify-valid is
structural. Under gold-differential execution on the 180 spec prompts,
63.2\% (arith+func) and 36.4\% (linalg) of verify-valid generations
compute the correct function under the canonical interface
(Table~\ref{tab:verify_vs_functional}). Much of the linalg shortfall is
interface non-conformance, a diagnosis supported most directly by the
$n{=}10$ adapted-wrapper experiment. A signature-conformance constraint
layer (provisional $C_4$) and curated growth of the functional reference
set are the headline open problems (\S\ref{sec:future}).

\paragraph{arith+func is parity-to-loss; StableHLO is
corpus-conditional.} Across three runs
(Table~\ref{tab:multiseed}), linalg is robust (all three runs at
80.0\%; $+21$--$44$pp over baselines with non-overlapping
CIs), arith+func is parity-to-loss (within CI of the 34B baselines,
trailing StarCoder2-15B by 13.6pp non-overlapping), and StableHLO
holds only on Spec-30 (baseline saturation on Held-Out-200,
\S\ref{sec:results} Table~\ref{tab:stablehlo_cross_system}). The
headline is dialect- and corpus-specific.

\paragraph{Single SLM family, single hardware, single version pin.}
Primary results use SmolLM2-1.7B-Instruct (Phi-3.5-mini appears only
in the \S\ref{sec:ablations} HCS replication); we do not sweep the
broader SLM landscape. All numbers are collected on a single Apple M4
Max laptop with SmolLM2 at fp16 and the 15B/34B baselines at 4-bit
quantization (Q4\_K\_M; StarCoder2 Q4\_0); reproducibility on CUDA,
CPU-only, or larger Apple Silicon is untested. The verifier toolchain
is pinned to LLVM \texttt{llvmorg-19.1.7} and \texttt{iree-compiler
20241104.1068}. Drift is characterized for the StableHLO/iree path: all
230 in-scope released StableHLO modules verify identically under the
pinned iree-compiler and under iree-base-compiler 3.11.0rc (about 16
months later), the only divergence being 1 of the 25 out-of-grammar
stress programs, outside every reported metric. Drift on the LLVM
\texttt{mlir-opt} path used by arith+func and linalg is not
characterized.

\paragraph{Quantization asymmetry, controlled.} A same-family
Granite-Code-8B-fp16 control beats its own family's 34B Q4 baseline on
both dialects ($+16.7$pp arith+func, $+14.4$pp linalg) despite
$4\times$ fewer parameters, so quantization rather than parameter count
binds for the Granite family. The linalg lead survives (the 8B-fp16
control still trails SmolLM2 by 29.9pp); on arith+func it beats SmolLM2
by $+15.0$pp paired ($p<0.001$), strengthening the non-win scoping. A
34B-fp16 cell (68+ GB of weights) does not fit the laptop budget.

\paragraph{Grammar coverage.} The released grammars cover
arith+func+memref, linalg (12 named ops), and StableHLO core
(${\sim}40$ ops); other dialects (scf, affine, gpu, async, transform,
tensor, vector) are not covered, though the same
\texttt{llvm-tblgen}$\to$grammar pipeline applies. Out-of-grammar
prompts return no output rather than malformed IR
(\S\ref{sec:results}).

\paragraph{In-line coupled decoder.} Theorem~\ref{thm:eq} establishes
equivalence to post-hoc rejection sampling, but our attempted in-line
implementation reaches \texttt{max\_tokens} before grammar accept on
every prompt; all reported numbers use post-hoc rejection sampling
(\S\ref{sec:future}).

\section{Future Work}
\label{sec:future}

The most consequential gap is the structural-vs-functional
distinction (\S\ref{sec:limits}). The released gold-differential
harness already executes generations against the gold module of all
180 spec prompts on randomized inputs
(Table~\ref{tab:verify_vs_functional}). The next step is curated growth
of the hand-authored functional reference set, with reference
implementations and property-based comparators for shape-polymorphic
prompts, as a versioned release; we did not change the released data
during review.

A second thread is $C_4$, a shape- and interface-aware constraint
layer. Two observations motivate it: 13 of the 22 verify-valid linalg
generations in the gold-differential run declare a non-conforming
interface, and the per-op linalg breakdown (Fig.~\ref{fig:perop}) shows SmolLM2 +
$C_1$+$C_2$+$C_3$ at 100\% on most named ops (\texttt{add},
\texttt{sub}, \texttt{mul}, \texttt{div}, \texttt{abs}, \texttt{copy},
\texttt{matmul}, \texttt{matvec}) but trailing on shape-reasoning ops
(\texttt{transpose} 4/20, \texttt{broadcast} 0/2) and on \texttt{fill}
and \texttt{exp} (1/5 and 1/2),
where the residual ``other'' error bucket (Table~\ref{tab:err}) is
dominated by rank/shape mismatches that pass type-domain and
SSA-scope checks but fail the verifier's shape-inference pass. $C_4$
would condition the next operand's tensor type on the partial
program's shape function (declared in ODS via
\texttt{InferShapedTypeOpInterface}, extractable by the same
\texttt{tblgen} pipeline as $C_2$); we have a sketch but not an
implementation, and identify $C_4$ as the most direct way to push
the linalg ceiling above 80.0\%.

Two smaller threads: (i) the in-line coupled decoder hit
\texttt{max\_tokens} on every prompt in our pilot
(\S\ref{sec:limits}), leaving the empirical-decoder gap as future
work for anyone aiming to remove the 5-retry budget; (ii) extending
the grammar to scf, affine, gpu, async, transform, tensor, vector is
a few-day derivation per dialect on the existing pipeline, not a
research question.

\section{Conclusion}
\label{sec:conclusion}

On dialects whose verifier semantics are dominated by structural
constraints, a schema-derived three-layer constraint stack lifts a
1.7B-parameter SLM, as a system with token-level enforcement, to
verify-valid rates that match or exceed 15B/34B open-weight code
baselines running post-hoc rejection on linalg (mean of three runs
80.0\%, $+21.3$ to $+44.3$ pp over baselines with non-overlapping CIs,
surviving a same-family fp16 precision control by 29.9 pp) and on
hand-curated StableHLO-Spec-30 (63.3\%, $+26.6$ to $+30.0$ pp over the
34B baselines, within CI of StarCoder2-15B). The StableHLO win is
corpus-conditional: the same baselines saturate at 98--100\% on the
templated parametric Held-Out-200 (61.5\% SmolLM2), so we retain the
lead only on Spec-30. On arith+func the result is
parity-to-loss, which we scope as the boundary of training-free
constrained decoding rather than a uniformity claim. Verify-valid remains
a structural measure: 36.4\% of verify-valid linalg generations compute
the correct function under the canonical interface, so closing the
functional gap is the main open problem. The construction
is mechanical: every $C_i$ is derived from MLIR ODS through one
tooling chain, so the procedure applies unchanged to any future
dialect that ships the schema. We release four NL$\to$MLIR benchmarks,
the StableHLO-OutOfGrammar-25 stress set, the MLIR-Functional-Reference-30
set, and the gold-differential harness under Apache-2.0 with Gebru
datasheets, Croissant metadata, and a tagged code release that carries
every per-prompt generation.

\label{cr:mainend}

\section*{Acknowledgments and Disclosure of Funding}
\paragraph{Funding.} This work received no third-party funding or support.
All experiments ran on a single Apple M4 Max laptop owned by the author; no
grants, cloud credits, or hardware donations were used.
\paragraph{Competing interests.} The author is employed by Meta, the
developer of Code Llama, one of the baseline models evaluated in this paper.
The work was conducted in the author's personal capacity, outside the scope
of that employment, and Meta had no role in its design, execution, analysis,
or the decision to publish. The author reports no other competing interests.

\bibliographystyle{plain}
\bibliography{refs}

\appendix
\section*{Appendix}

\section{Soundness and coverage of the coupled decoder}
\label{app:soundness}

We formalize the two key properties of the in-line C3 coupled decoder
(\S\ref{sec:method}): (i) every generation it accepts passes the post-hoc
scope validator, and (ii) every generation passing the scope validator is
reachable by some token-oracle trajectory.

\subsection*{Definitions}

Let $\mathcal{G}$ denote the LALR-compiled generation grammar over terminal
alphabet $\Sigma$, and let $T$ denote the set of terminals partitioned into
literal and regex-matched classes. Let $V$ denote the BPE token vocabulary
of the language model. Each decode step produces $v \in V$; $\mathrm{text}(v)$
denotes the vocab string.

A \emph{decoder state} $\sigma = (q, \Lambda, \phi, p)$ is a tuple of:
\begin{itemize}
  \item $q$: an LALR interactive-parser state (position in $\mathcal{G}$'s
  item-set lattice).
  \item $\Lambda$: a symbol table mapping SSA names to type stubs.
  \item $\phi \in \Phi = \{\textsc{Outer, Param, FuncBody, OpStart, AfterEq,
  RetVal, Closed}\}$: an op-phase marker.
  \item $p \in \Sigma^*$: the accumulated partial text since the last
  closed terminal.
\end{itemize}

The decoder's token-level mask at state $\sigma$ is
\[
\mathrm{mask}(\sigma) = \bigcup_{t \in \mathrm{legal}(q)}
    \mathrm{prefix}(t, p) \cap \mathrm{scope}(t, \phi, \Lambda)
\]
where $\mathrm{legal}(q)$ enumerates shift-legal terminals at parser state
$q$, $\mathrm{prefix}(t, p)$ is the set of $v \in V$ whose text extends $p$
toward a valid completion of terminal $t$, and $\mathrm{scope}(t, \phi,
\Lambda)$ narrows the set for SSA-typed terminals based on the scope
predicate:

\[
\mathrm{scope}(t, \phi, \Lambda) =
\begin{cases}
    V & t \not\in \{\mathrm{SSA}\} \\
    \mathrm{prefix}(\Lambda, p) & t = \mathrm{SSA}, \phi \in \{\textsc{AfterEq, RetVal}\} \\
    \mathrm{prefix}(\Sigma_{\mathrm{SSA}}, p) \setminus \mathrm{prefix}(\Lambda, p)
    & t = \mathrm{SSA}, \phi \in \{\textsc{Param, OpStart}\} \\
    V & \text{otherwise.}
\end{cases}
\]
$\Sigma_{\mathrm{SSA}}$ denotes the SSA regex character class.

State transitions $\sigma \xrightarrow{v} \sigma'$ are defined operationally:
append $\mathrm{text}(v)$ to $p$; greedily close zero or more terminals whose
regex fullmatches the accumulated $p$; each closure advances the LALR state,
possibly updates $\Lambda$ (on DEF) or checks $\Lambda$ (on USE), and
transitions $\phi$ per the phase table.

\subsection*{Theorem 1 (Soundness)}
\thmanchor{1}{thm:sound}
If the decoder reaches a final state $\sigma_f = (q_f, \Lambda_f, \phi_f, p_f)$
with $q_f$ an accept state and $p_f = \varepsilon$, then the emitted string
$s$ satisfies:
\begin{enumerate}
  \item $s$ parses under $\mathcal{G}$ (and therefore under $\mathrm{mlir.lark}$,
  since $\mathcal{G}$ is a subset by grammar inclusion).
  \item The post-hoc validator \texttt{c3\_scope.validate(s)} returns
  accept.
\end{enumerate}

\paragraph{Proof sketch.} (i) follows from LALR closure: the interactive parser
guarantees $q_f$ accepts iff the consumed terminal sequence is a
$\mathcal{G}$-derivation of $s$, and the reduce actions that $q_f$ must have
taken are exactly the leftmost derivation steps.

(ii) The symbol-table $\Lambda$ is updated monotonically at each SSA DEF
(on $\phi \in \{\textsc{Param, OpStart} \cup \textsc{AfterEq-transition}\}$)
and checked at each SSA USE ($\phi \in \{\textsc{AfterEq, RetVal}\}$).
Because the mask $\mathrm{scope}(\mathrm{SSA}, \phi, \Lambda)$ restricts USE
tokens to $\mathrm{prefix}(\Lambda, p)$, any USE that closes does so with a
name in $\Lambda$. Therefore $s$ has no out-of-scope SSA references: every
use is matched by a prior definition in the same function. This is the
definition of $\texttt{c3\_scope}$ accept.\proofend

\subsection*{Theorem 2 (Coverage, BPE-refined)}
\thmanchor{2}{thm:cov}
For any MLIR fragment $s^*$ that parses under $\mathcal{G}$ \emph{and}
passes \texttt{c3\_scope.validate}, there exists an oracle token sequence
$(v_1, \ldots, v_n) \in V^n$ such that the decoder state machine reaches a
final accept state with emitted string $s^*$.

We strengthen the earlier proof by handling BPE-boundary misalignment
explicitly. The subtlety is that the BPE tokenization of $s^*$ taken as a
whole need not respect terminal boundaries: a single token may decode
to text spanning two terminals (e.g., the SmolLM2 token
\texttt{module\textvisiblespace} carries MODULE plus a trailing WS), and the
per-terminal tokenization of a substring may not be a subsequence of the
whole-string BPE.

\paragraph{Proof.} We construct an oracle trajectory character-by-character
 and then show it factors into at least one BPE tokenization
compatible with the vocab.

Let $s^* = c_1 c_2 \cdots c_N$ (over the character alphabet) and let
$(t_1, \ldots, t_m)$ be the leftmost LALR-derivation terminals with text
$w_i$, so that $s^* = w_1 w_2 \cdots w_m$. Define the \emph{character
oracle} as a sequence of single-character vocab tokens. Our vocabulary
$V$ covers all printable ASCII characters individually (a standard
property, true for all BPE tokenizers considered in \S\ref{sec:exp}). Feeding $c_1, \ldots, c_N$ through the decoder advances $p$
one character at a time; at each character boundary, the close-terminals
procedure commits a terminal iff the accumulated $p$ is a maximal match
for some legal terminal. Because the LALR derivation is unambiguous over
$\mathcal{G}$, the closures fire exactly at the boundaries between
$w_i$ and $w_{i+1}$, so the decoder advances the parser state in
lockstep with the leftmost derivation. At each SSA USE step, the
scope-predicate $\mathrm{scope}(\mathrm{SSA}, \phi, \Lambda)$ requires
the next token's text to start some name in $\Lambda$; because $s^*$
passes \texttt{c3\_scope}, that name is present by construction, and
the single-character token $c$ at the start of a SSA-USE terminal is a
prefix of some name in $\Lambda$. Therefore the character oracle
reaches an accept state.

To lift this to the whole-string BPE setting, observe that if the
character oracle is accepted, any \emph{coarser} tokenization (that is, a
tokenization where some consecutive characters are merged into a
multi-character token $v$ such that $\mathrm{text}(v) = c_i c_{i+1}
\cdots c_j$ and $v \in V$) is also accepted, \emph{provided} the
decoder's per-step mask admits $v$ as a legal next token at the
corresponding state. The multi-character case is handled by
``carry-over'': the decoder's close-terminals loop consumes complete
terminals from the start of $p$ and preserves leftover text as the new
$p$, so a token whose text spans two terminals (say $w_i w_{i+1}'$ with
$w_{i+1}'$ a prefix of $w_{i+1}$) results in $w_i$ being closed and
$w_{i+1}'$ remaining as the leftover $p$. The per-step mask under the
resulting post-close state must admit tokens that extend $p$ toward
$w_{i+1}$; this is guaranteed by the two-component mask construction
(\S\ref{sec:method} and the earlier $\mathrm{mask}(\sigma)$ definition
in this appendix), which unions
(a)~$\mathrm{prefix}(t', p)$ for each legal $t'$ under the
post-close state and (b)~the \emph{lookahead} contribution: if $p$
completely matches some post-close legal terminal, the mask also
includes $\mathrm{prefix}(t'', \varepsilon)$ for $t''$ legal in the
\emph{next} state. Therefore any whole-string tokenization of $s^*$
that is factorizable into vocab tokens is a valid oracle.\proofend

\paragraph{On the ``vocab-covers-ASCII'' assumption.} All tokenizers
used in the paper (SmolLM2/Phi family, StarCoder2, CodeLlama, Granite)
include single-character tokens for every printable ASCII character in
their vocabularies. If a tokenizer \emph{did not} include some single
character $c$ used by $\mathcal{G}$'s terminal alphabet, the
character-oracle step of the proof would fail and coverage would be
subject to tokenizer-specific conditions. We do not observe this
failure mode in practice; the lemma as stated suffices for all
configurations reported in \S\ref{sec:results}.

\subsection*{Theorem 3 (Equivalence to post-hoc rejection sampling)}
\thmanchor{3}{thm:eq}
Let $D_{\mathrm{in}}$ be the in-line coupled decoder and $D_{\mathrm{post}}$
be the post-hoc rejection-sampled decoder (generate under C1+C2 freely;
accept iff parse + \texttt{c3\_scope.validate}). Both decoders induce the
same set of \emph{acceptable} output strings:
\[
\mathrm{Accept}(D_{\mathrm{in}}) = \mathrm{Accept}(D_{\mathrm{post}}).
\]

\paragraph{Proof sketch.} Forward inclusion is Theorem 1. Reverse inclusion
is Theorem 2: every $s^*$ accepted by $D_{\mathrm{post}}$ is reachable by
$D_{\mathrm{in}}$ via the oracle trajectory built from the BPE tokenization
of $s^*$, so $s^* \in \mathrm{Accept}(D_{\mathrm{in}})$.\proofend

\subsection*{Remark: efficiency}
Soundness + coverage ensure correctness parity, but the decoders differ in
sampling efficiency: $D_{\mathrm{in}}$ achieves 1 sample per output, while
$D_{\mathrm{post}}$ requires an expected $1/\mathrm{Pr}[\text{accept}]$
samples. On arith+func with SmolLM2 and a 5-retry budget, the
observed mean attempts for $D_{\mathrm{post}}$ is 1.83 (Day 5); the in-line
decoder reduces this to 1.0 deterministically. The theorems above do not
bound sampling quality (temperature, mode collapse), only the reachable
support.

\subsection*{Remark: agreement between C3 and \texttt{mlir-opt} is empirical}
Theorem~\ref{thm:sound} is stated in terms of agreement between the in-line decoder
and the \texttt{c3\_scope} post-hoc filter. Whether \texttt{c3\_scope}
agrees with the downstream \texttt{mlir-opt} verifier is a separate,
empirical property, and it is not perfect. We measure it directly on
the frozen released pools by applying \texttt{c3\_scope} to every
parse-valid output of the $C_1$ rung (and, for SmolLM2, also the
$C_1{+}C_2$ rung) and cross-tabulating against the recorded
verify outcome. A false reject is a program that \texttt{mlir-opt}
accepts and \texttt{c3\_scope} rejects. Table~\ref{tab:c3_false_rejects}
reports the counts.

\begin{table}[h]
\centering
\caption{\texttt{c3\_scope} false rejects against \texttt{mlir-opt} on
parse-valid generations from the frozen released pools (arith+func
$n{=}200$, linalg $n{=}125$). Each cell is false rejects / verify-valid
programs. Every baseline cell is zero, so the shared $C_3$ gate never
discards a program the verifier accepts; the few SmolLM2 false rejects
come from multi-line operations, as the text explains.}
\label{tab:c3_false_rejects}
\small
\begin{tabular}{lrr}
\toprule
Generations & arith+func & linalg \\
\midrule
SmolLM2-1.7B ($C_1$ rung) & 3 / 89 & 1 / 80 \\
SmolLM2-1.7B ($C_1{+}C_2$ rung) & 5 / 91 & 1 / 92 \\
CodeLlama-34B & 0 / 121 & 0 / 71 \\
StarCoder2-15B & 0 / 134 & 0 / 59 \\
Granite-Code-34B & 0 / 94 & 0 / 42 \\
\bottomrule
\end{tabular}
\end{table}

The filter never rejects a valid baseline program, so the shared $C_3$ gate
does not penalize the 15B/34B systems. The few SmolLM2 false rejects all
share one cause: the grammar allows an operation to span several lines
(e.g., \mbox{\texttt{arith.extf \%x :\ f16}} followed by \texttt{to f64} on the
next line), and the line-based scope tracker then reports an undefined
use (9 cases) or a type mismatch (2 cases). This is a parser limitation
of \texttt{c3\_scope}, not an error in the theorems; normalizing
multi-line operations before scope tracking would remove it. The net
effect of $C_3$ on SmolLM2 remains positive ($+6.5$pp on arith+func,
\S\ref{sec:ablations}) because it rejects far more invalid programs
than valid ones (Fig.~\ref{fig:scope}).

\subsection*{Empirical gap of the in-line construction}
We implement the in-line decoder with 82 unit tests and end-to-end
parse-valid MLIR on a canonical prompt, and release it alongside the
rejection sampler. On a paired $n{=}10$ test against rejection sampling,
the greedy in-line variant reaches parse-valid on only 20\% of prompts
and runs roughly $44\times$ slower per generation; the remaining 80\%
of runs hit \texttt{max\_tokens} before the grammar reaches its accept
state. Temperature-annealed retries did not close the gap. The gap is
a sampling-coverage issue rather than a soundness issue: accepted
strings remain a subset of those admitted by post-hoc rejection
sampling, consistent with Theorem~\ref{thm:eq}.

\paragraph{Diagnosis.} The per-step mask is the intersection of C1's
shift-legal terminals and C3's in-scope name trie, which has nonzero
probability of becoming empty after a long prefix has committed to a
partial BPE token that no in-scope name can complete. In those states
the decoder is forced into low-probability tail tokens to recover, and
the resulting sequences exceed \texttt{max\_tokens} before reaching
an accept state. A future variant that backtracks past the BPE-boundary
commit point rather than sampling forward from it would close the gap;
we leave this to future work and report all empirical results from the
post-hoc form.

\section{Notation and naming}
\label{app:notation}

\begin{center}
\small
\begin{tabularx}{\linewidth}{@{}lL@{}}
\toprule
Term & Meaning \\
\midrule
$C_1$ & Context-free grammar over the target dialect's operation signatures, derived from ODS; a token-level mask for SmolLM2, a post-hoc parse check for the baselines \\
$C_2$ & Type-domain splits of $C_1$ read from the ODS type lattice (token-level mask) \\
$C_3$ & SSA-scope validator; drives five-retry rejection sampling \\
free & One unconstrained attempt \\
verify-valid & The dialect verifier accepts the program (\texttt{mlir-opt --verify-diagnostics}, or \texttt{iree-compile --compile-to=input} for StableHLO); structural, not functional \\
output-match & A generation's output matches the reference on a fixed input (MLIR-Functional-Reference-30) \\
gold-differential & A generation matches the benchmark's gold module on five seeded random input draws, after a gold-vs-gold determinism gate \\
evaluation pool & The frozen arith+func ($n{=}200$) and linalg ($n{=}125$) prompt sets: spec prompts plus padding prompts (\S\ref{sec:exp}) \\
run & One pass over a pool; three runs use seeds 0, 1, 2 (repeats for SmolLM2, whose retries are unseeded) \\
converse cell & SmolLM2 evaluated under the baselines' post-hoc protocol \\
best configuration & A baseline's higher-scoring setting of $C_1$-only and $C_1{+}C_3$ \\
\midrule
MLIR-Spec-150 & 150 arith+func+memref pairs \\
Linalg-Spec-30 & 30 linalg (memref) pairs \\
StableHLO-Spec-30 & 30 hand-curated StableHLO pairs \\
StableHLO-Held-Out-200 & 200 parametric-sweep StableHLO programs \\
StableHLO-OutOfGrammar-25 & 25 StableHLO programs outside the grammar (stress set) \\
MLIR-Functional-Reference-30 & Hand-authored input and expected-output references for 30 Spec prompts, 10 per dialect (no new prompts) \\
\bottomrule
\end{tabularx}
\end{center}

\section{Reproducibility, Datasheet, and Responsible Release}
\label{sec:reproducibility}

\paragraph{Artifacts.} Everything is released in two places. The
public repository at
\url{https://github.com/plawanrath/slm-to-mlir-constrained-emitter}
holds the code, benchmarks, and documentation; its tagged release
\texttt{v1.0-camera-ready} (\url{https://github.com/plawanrath/slm-to-mlir-constrained-emitter/releases/tag/v1.0-camera-ready})
freezes the camera-ready state and attaches every per-prompt generation
behind a number, table, or figure as \texttt{paper\_results.tar.gz}.
The benchmarks are also mirrored as Hugging Face Datasets in the
collection
\url{https://huggingface.co/collections/plawanrath/cross-dialect-mlir-benchmarks-neurips-2026-e-and-d}.
The two hold different things. The Hugging Face datasets are the
benchmarks themselves: prompts, gold modules, functional references,
and their metadata. The release asset holds model outputs on those
benchmarks: one row per prompt for every system and cell, with the
generated program, its parse and verify outcomes, and the attempt count,
plus the aggregate files. Every reported rate can be recomputed from it
without rerunning inference; it is supporting evidence, not a dataset
contribution. Together they contain: (i) the four
primary benchmarks plus the StableHLO-OutOfGrammar-25 stress set,
totaling 435 instances ($150 + 30 + 30 + 200 + 25$), and the hand-authored
functional reference set at \texttt{eval/functional/references.json}
($n{=}30$, 10 Spec prompts per dialect drawn from the three Spec
benchmarks, so not counted again in the 435; released as an evaluation artifact
for bounding the verify-valid $\to$ functional gap, not as a training
or fine-tuning set) with the gold-differential harness
(\texttt{eval/functional\_differential/}) and its per-prompt results for
all 180 spec prompts (in the release asset); (ii) the constrained-decoder source with frozen
\texttt{requirements.lock.txt}; (iii) the verifier wrappers
(\texttt{scripts/env/verify\_cache.py} for \texttt{mlir-opt},
\texttt{scripts/env/verify\_stablehlo.py}); (iv) seeded bootstrap
scripts and the scripts that regenerate every figure and table from the
raw per-prompt \texttt{.jsonl} rows in the release asset (the repository README maps each
table and figure to its script); (v)
per-benchmark Croissant records at
\texttt{eval/benchmarks/<name>/croissant.json} and a single Gebru-style
datasheet at \texttt{docs/datasheets/datasheet.md}. License is
Apache-2.0 (SPDX: Apache-2.0) for both code and data. The benchmarks
are intended for evaluation only. To prevent benchmark contamination
we ask users not to fine-tune on them; this is a stated intended-use
request, not a license term, since Apache-2.0 does not restrict
training. The request is recorded in the datasheet's ``uses not
supported'' field and in the Croissant \texttt{rai:dataLimitations}
field of each benchmark.

\paragraph{Hosted benchmark URLs.} The six dataset repositories are
grouped in the Hugging Face collection \url{https://huggingface.co/collections/plawanrath/cross-dialect-mlir-benchmarks-neurips-2026-e-and-d}:
\begin{itemize}
\item \url{https://huggingface.co/datasets/plawanrath/MLIR-Spec-150}
\item \url{https://huggingface.co/datasets/plawanrath/Linalg-Spec-30}
\item \url{https://huggingface.co/datasets/plawanrath/StableHLO-Spec-30}
\item \url{https://huggingface.co/datasets/plawanrath/StableHLO-Held-Out-200}
\item \url{https://huggingface.co/datasets/plawanrath/StableHLO-OutOfGrammar-25}
\item \url{https://huggingface.co/datasets/plawanrath/MLIR-Functional-Reference-30}
\end{itemize}
Each ships its Croissant 1.1 record (core and Responsible-AI fields,
with provenance) and per-instance JSON schema; the
HuggingFace copies are byte-identical to the JSONL files in the
repository (\texttt{eval/benchmarks/<name>/}).

\paragraph{Reproducibility pins.} All numbers are pinned to LLVM
\texttt{llvmorg-19.1.7} and iree-compiler 20241104.1068; both pins live
in \texttt{scripts/env/requirements.lock.txt} along with every other
host-side dependency at exact versions. The paired bootstrap (10,000
resamples, 95\% percentile CI) is deterministic under
\texttt{numpy.random.default\_rng(0)}, and every table and figure is
regenerated from the released per-prompt \texttt{.jsonl} files by the
scripts the repository README lists. Baseline generations reproduce
byte-identically under Ollama 0.32.1. One verifier gotcha worth
flagging up-front:
\texttt{iree-compile --compile-to=input} accepts empty stdin as a valid
empty \texttt{module \{\}} and returns success, which means a
constrained decoder that emits the empty string is silently scored as
verify-valid. Our wrapper \texttt{scripts/env/verify\_stablehlo.py}
requires a \texttt{func.func @} substring in the input before invoking
\texttt{iree-compile}. We recommend any reuser of the StableHLO
benchmarks apply the same guard. The verifier-tool substitution itself
is independently validated: \texttt{iree-compile --compile-to=input}
and upstream \texttt{stablehlo-opt v1.4.0} agree on 50/50 instances on
a stratified $n{=}50$ sample across six op families (build commands and
per-instance log in Appendix~\ref{app:stablehlo_toolchain}), so reusers comparing
against an in-tree \texttt{stablehlo-opt} environment should not see
material drift.

\paragraph{Datasheet summary.} Each instance is a (natural-language
description, reference MLIR program, dialect, difficulty tag, source
provenance) tuple. All reference MLIR programs are verifier-clean at
release time. Spec-* benchmarks were author-curated against public ODS
specs and filtered through the verifier; Held-Out-200 was generated by
a parametric sweep (585 raw candidates to 200 verifier-clean kept,
sweeping 7 op families $\times$ 6 dtypes $\times$ 3 shape ranks);
OutOfGrammar-25 was hand-authored against StableHLO ops the grammar
does not cover. There is no human-subject data, no PII, no crowdsourced
labor, no offensive content; the data domain is technical MLIR code
and English descriptions of numerical operations. The datasheet
(\S\ref{sec:datasheet}) covers every Gebru-template field including
the explicit ``uses the dataset does not support'' (fine-tuning, and
functional-correctness claims from verify-valid alone) for reuser
clarity. Splits: not
pre-split. All instances are released as a test set; users running
methods that require a training set are expected to hold programs out
themselves.

\paragraph{Role of the verify-valid metric.} We adopt verify-valid pass
rate as the primary metric for three reasons. First, it is mechanical:
no human judgment is required, so the benchmarks are reproducible
without grader drift. Second, it is well-defined per dialect and each
MLIR dialect ships its own verifier in-tree, so the metric extends to
any future dialect with no methodological redesign. Third, it is a
necessary condition for downstream usefulness: a program that fails the
verifier cannot be lowered, optimized, or executed regardless of any
other property. We deliberately do not adopt soft or token-level
metrics (BLEU, exact-match) because they conflate stylistic differences
with semantic ones in the IR setting, where two textually different
programs can be equivalent (e.g., differing only in SSA-name choice)
and two textually similar programs can differ in verify status (e.g.,
a single mismatched dtype).

\paragraph{Ethical considerations.} The benchmark task is technical IR
generation; we do not see direct dual-use risk from a verify-valid
evaluation. Two indirect considerations are worth recording. (i) The
pretraining corpora of all evaluated models (SmolLM2, CodeLlama,
Granite-Code, StarCoder2) include substantial open-source code, and
there is no auditable line between ``MLIR seen during pretraining'' and
``verify-valid-cleared MLIR generated at inference''; reviewers should
read absolute verify rates as artifacts of the pretraining-plus-constraint
-stack composition rather than as model-intrinsic capability
measurements. (ii) We evaluate only limited open-weight models. We
disclose this as a deliberate scope-of-evaluation rather than an
ethical claim about the excluded families; a future replication that
includes them requires no methodological change, only additional
compute.

\section{StableHLO toolchain and verifier substitution}
\label{app:stablehlo_toolchain}

The canonical structural verifier for StableHLO is \texttt{stablehlo-opt}
from the in-tree \texttt{openxla/stablehlo} repository. We use
\texttt{iree-compile --iree-input-type=stablehlo --compile-to=input}~\cite{iree2024}
as the verifier in our pipeline; it runs StableHLO's in-tree
verification (the same \texttt{mlir::stablehlo::registerAllDialects}
and \texttt{verify()} machinery \texttt{stablehlo-opt} invokes) as the
first stage of the IREE compile pipeline and exits before any
lowering. We use \texttt{iree-compile} rather than
\texttt{stablehlo-opt} throughout the body of the paper because the
vendored \texttt{mlir-opt} in our reproducibility container does not
ship the StableHLO dialect.

\paragraph{Verifier-tool concordance ($n{=}50$).} To validate the
substitution against the canonical verifier quantitatively, we built
\texttt{stablehlo-opt v1.4.0} from source against our reproducibility
container's LLVM 19.1.7 (\texttt{openxla/stablehlo} HEAD requires a
more recent LLVM than our pin; v1.4.0 at LLVM SHA
\texttt{9ddfe62f5c11} configures cleanly and builds in $\sim$3 min
via Ninja) and ran both verifiers on a stratified $n{=}50$ sample
drawn from \texttt{StableHLO-Held-Out-200} $\cup$
\texttt{StableHLO-Spec-30}, covering six op families
(elementwise-binary 26, transpose 6, dot\_general 4, broadcast 3,
reshape 3, other 8). \textbf{Result: 50/50 concordant accepts,
0/50 \texttt{iree-compile}-accepts-\texttt{stablehlo-opt}-rejects,
0/50 reverse direction.} The substitution is empirically sound on the
released benchmarks. Build commands and the per-instance log are in
\texttt{scripts/verify\_concordance.py} and
\texttt{results/day51/verify\_concordance\_n50.json}.

\paragraph{Build recipe (for readers reproducing the canonical
verifier locally).} The following commands build \texttt{stablehlo-opt}
at the LLVM revision pinned by the StableHLO CI.

\begin{lstlisting}[basicstyle=\small\ttfamily,frame=single,breaklines=true]
git clone https://github.com/openxla/stablehlo.git && cd stablehlo
git submodule update --init --recursive
# pin LLVM revision tested by openxla/stablehlo CI:
LLVM_COMMIT=$(cat build_tools/llvm_version.txt)
cmake -GNinja -B build -DCMAKE_BUILD_TYPE=Release \
    -DLLVM_TARGETS_TO_BUILD="X86;ARM" \
    -DLLVM_EXTERNAL_PROJECTS=stablehlo \
    -DLLVM_EXTERNAL_STABLEHLO_SOURCE_DIR=$PWD
ninja -C build stablehlo-opt
\end{lstlisting}

Once built, \texttt{stablehlo-opt --verify-diagnostics input.mlir} is
a drop-in replacement for the IREE wrapper.

\paragraph{Grammar tooling caveats for reusers.}
Two reuser-facing caveats are worth recording, because the same
setup will reproduce both. First, four StableHLO grammar productions
(\texttt{op\_transpose}, \texttt{op\_broadcast\_in\_dim},
\texttt{op\_reshape}, \texttt{op\_dot\_general}) must avoid
\texttt{\textbackslash}-line-continuation tokens for splitting long
rules across source lines. LARK parses these correctly, but the
llguidance backend used by Outlines does not: it raises a silent
\texttt{LLMatcher} lexer error and emits zero tokens, producing 100\%
empty-string generations. The released grammars use single-line
productions; any reuser editing the grammars should verify that their
lark-parser and outlines+llguidance versions agree on every
multi-line rule before reporting verify rates. Second,
\texttt{iree-compile --iree-input-type=stablehlo --compile-to=input}
accepts empty stdin and returns success with stdout
\texttt{module \{\}}. This is a property of \texttt{iree-compile}'s
input handling, not a bug in our code, but it interacts
catastrophically with the first caveat: a constrained decoder that
produces empty output is silently scored as verify-valid. Symptom: a
``100\% verify-valid'' reading that is in fact ``100\% empty-string
accepted as empty module.'' Fix: our wrapper
\texttt{scripts/env/verify\_stablehlo.py} requires a
\texttt{func.func @} substring in the input before invoking
\texttt{iree-compile}; inputs that are empty or contain only an empty
module return a non-success error code. We recommend any reuser of the
StableHLO benchmarks adopt the same guard regardless of whether they
use Outlines.

\section{Extended figures and tables}
\label{app:extras}

\begin{figure}[h]
\centering
\includegraphics[width=\linewidth]{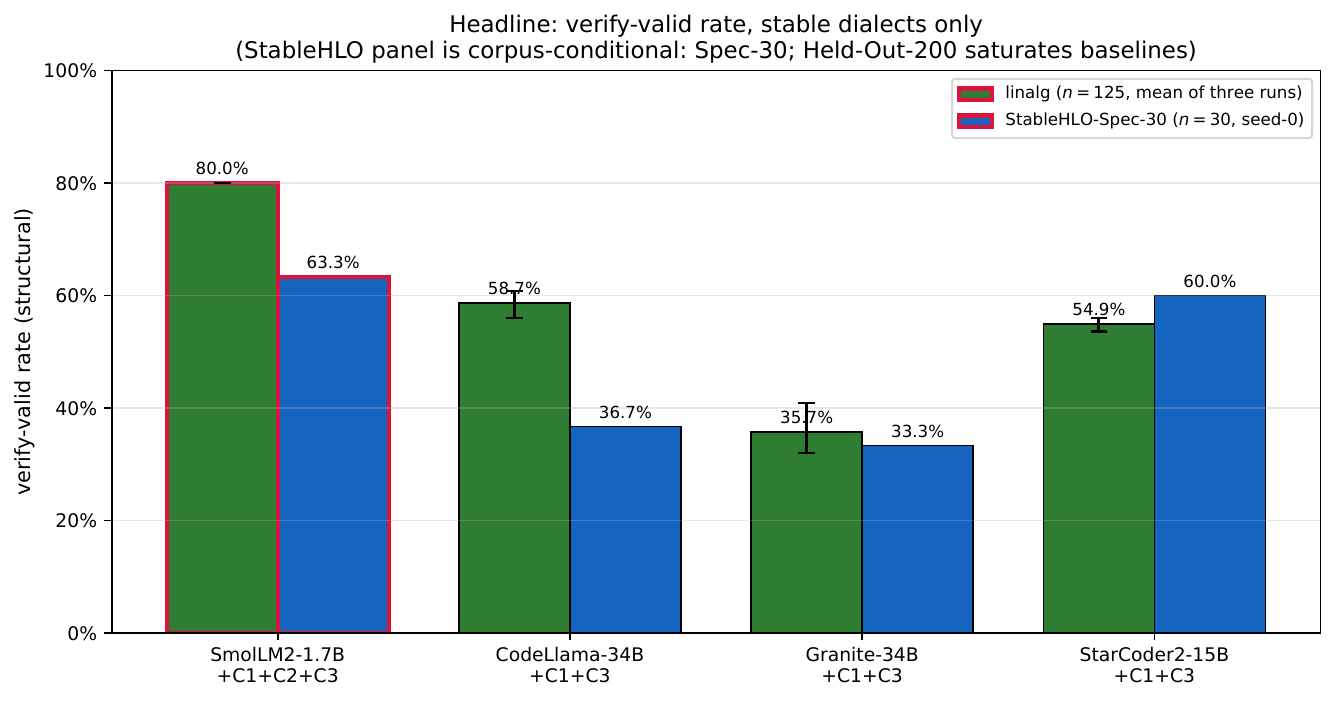}
\caption{Headline: \texttt{mlir-opt}/\texttt{iree-compile} verify-valid
rates on the two dialects where we report a win (linalg,
StableHLO). SmolLM2-1.7B + $C_1$+$C_2$+$C_3$ attains a higher rate
than each 15B/34B baseline under the same 5-retry rejection budget.
Linalg bars are means of three runs at $n{=}125$ with across-run
range whiskers; the SmolLM2 cell is 80.0\% (range 0pp). The StableHLO
panel is corpus-conditional and reports the seed-0 \emph{Spec-30}
corpus ($n{=}30$); the larger \emph{Held-Out-200} corpus saturates the
15B/34B baselines (verify rate $\approx$ 1.0) and is therefore
reported in a separate table rather than in the headline.
arith+func is intentionally omitted here and reported in
Table~\ref{tab:multiseed} as a non-win cell (mean of three runs 53.2\%
$\pm$ 1.8pp for SmolLM2 + $C_1$+$C_2$+$C_3$); ``verify-valid'' is a
structural property, not functional correctness (\S\ref{sec:limits}).}
\label{fig:main}
\end{figure}

\begin{figure}[h]
\centering
\includegraphics[width=0.85\linewidth]{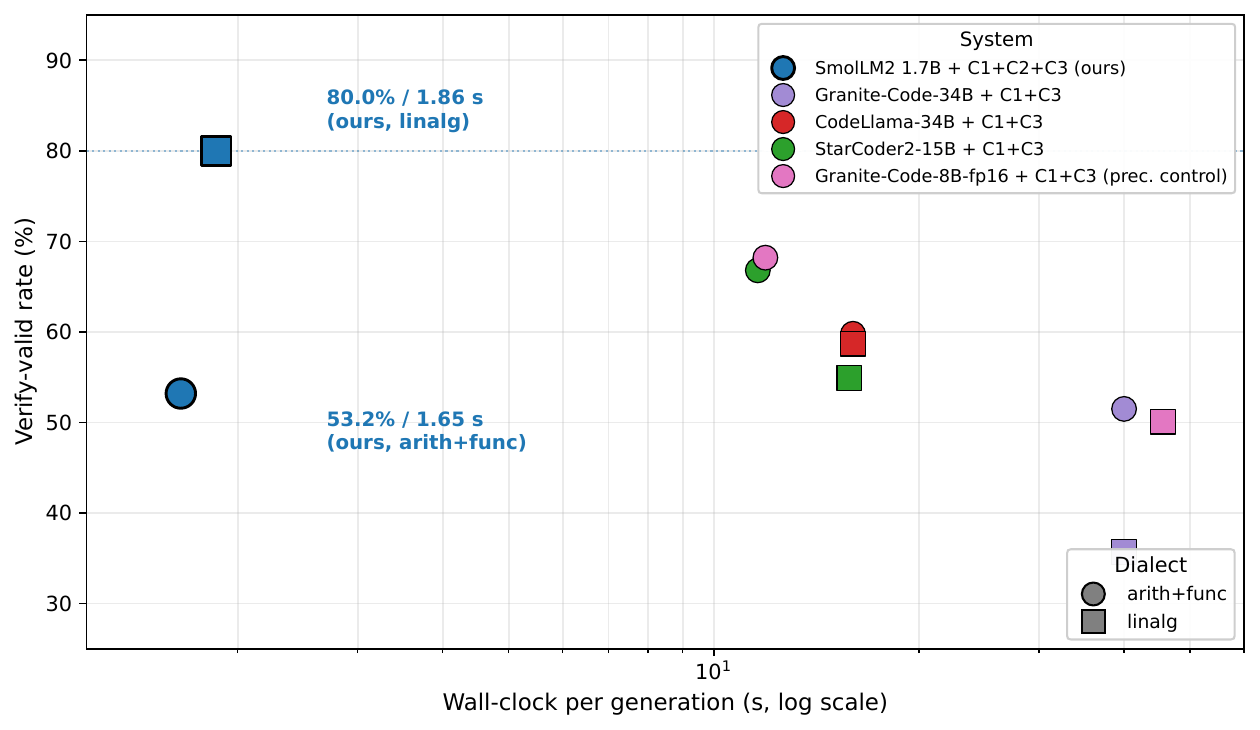}
\caption{Efficiency frontier: wall-clock per generation (log scale)
vs verify-valid rate (mean of three runs). Five systems
$\times$ two dialects, including a Granite-Code-8B-fp16 same-family
precision control. The SmolLM2 + $C_1$+$C_2$+$C_3$ linalg
point sits at 1.86\,s / 80.0\% (top-left, ours) and the
arith+func point at 1.65\,s / 53.2\%; on linalg the
SmolLM2 cell is Pareto-dominant by both axes against every other
cell. The 15B/34B 4-bit baselines run 6--24$\times$ slower per
generation; the Granite-Code-8B-fp16 control runs 7$\times$ slower on
arith+func and 25$\times$ slower on linalg. Color =
system, marker = dialect (circles = arith+func, squares = linalg).}
\label{fig:efficiency}
\end{figure}

\begin{figure}[h]
\centering
\includegraphics[width=0.95\linewidth]{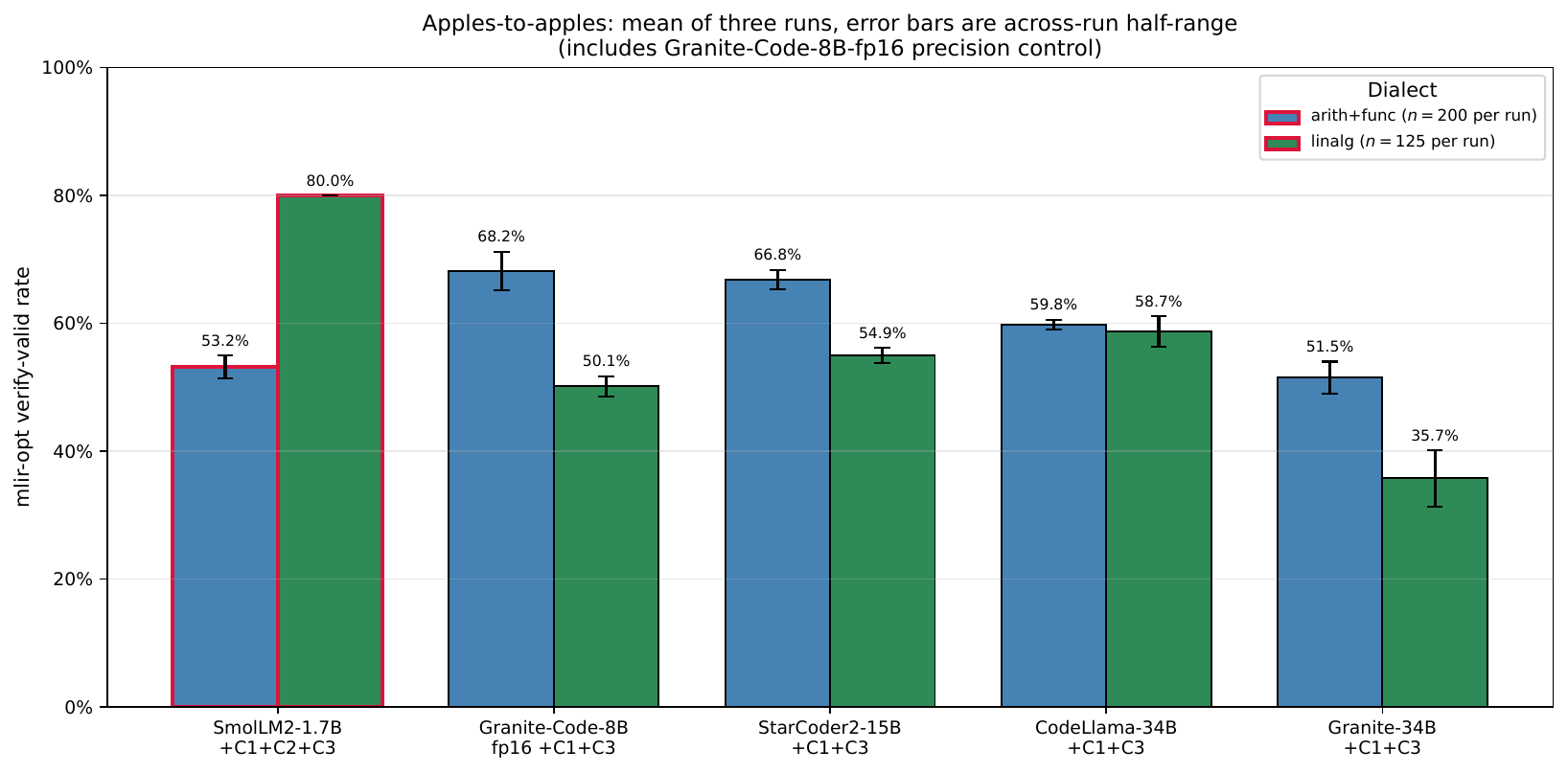}
\caption{Apples-to-apples comparison, mean of three runs with
across-run half-range whiskers. Five systems $\times$ two
dialects, including a Granite-Code-8B-fp16 precision control on
\emph{both} dialects. linalg: SmolLM2-1.7B +
$C_1$+$C_2$+$C_3$ (red border, ``ours'') = 80.0\%, leading
CodeLlama-34B (58.7\%) by $+21.3$pp, Granite-34B (35.7\%) by
$+44.3$pp, StarCoder2-15B (54.9\%) by $+25.1$pp, and the
Granite-Code-8B-fp16 control (50.1\% $\pm$ 1.6pp) by $+29.9$pp; CIs
non-overlapping across all five systems. arith+func: SmolLM2
(53.2\% $\pm$ 1.8pp) trails CodeLlama-34B (59.8\%) by $-6.6$pp (within
CI), narrowly leads Granite-34B (51.5\%) by $+1.7$pp (within CI),
trails StarCoder2-15B (66.8\%) by $-13.6$pp, and trails the
Granite-Code-8B-fp16 control (68.2\% $\pm$ 3.0pp) by $-15.0$pp.
$n{=}200$ per run for arith+func, $n{=}125$ per run for linalg,
three runs each.}
\label{fig:apples}
\end{figure}

\begin{figure}[h]
\centering
\includegraphics[width=\linewidth]{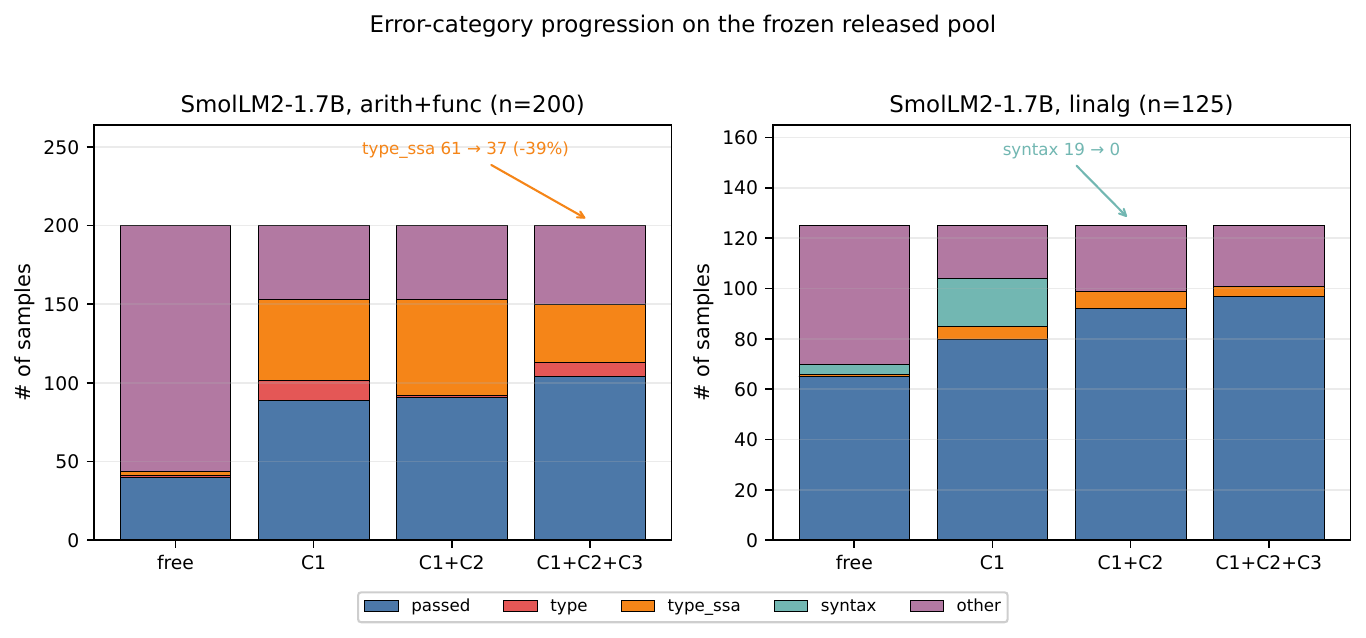}
\caption{Error-category progression across constraint levels,
SmolLM2-1.7B, frozen released pools. Each layer shrinks its target
bucket. arith+func: $C_3$ cuts \texttt{type\_ssa} errors $61 \to 37$
($-39\%$). linalg: $C_2$ eliminates \texttt{syntax} errors $19 \to 0$;
\texttt{type\_ssa} stays small (at most 7) because linalg's
\texttt{ins(...) outs(...)} re-declares types at each use site.}
\label{fig:err}
\end{figure}

\begin{figure}[h]
\centering
\includegraphics[width=0.95\linewidth]{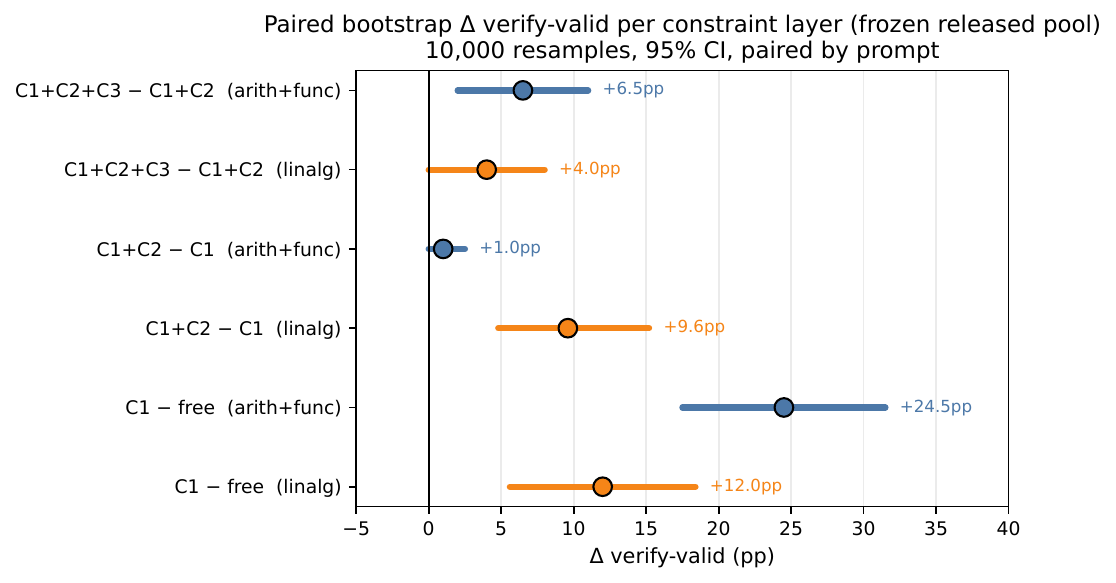}
\caption{Paired bootstrap $\Delta$verify per constraint layer with 95\%
CIs, frozen released pools. $C_1$+$C_2$+$C_3 - C_1$+$C_2$ is $+6.5$pp
on arith+func (CI $[+2.0, +11.0]$) and $+4.0$pp on linalg (CI $[+0.0,
+8.0]$); $C_1$+$C_2 - C_1$ is $+9.6$pp on linalg (CI $[+4.8, +15.2]$)
and $+1.0$pp on arith+func (CI $[+0.0, +2.5]$).}
\label{fig:paired}
\end{figure}

\begin{figure}[h]
\centering
\includegraphics[width=0.85\linewidth]{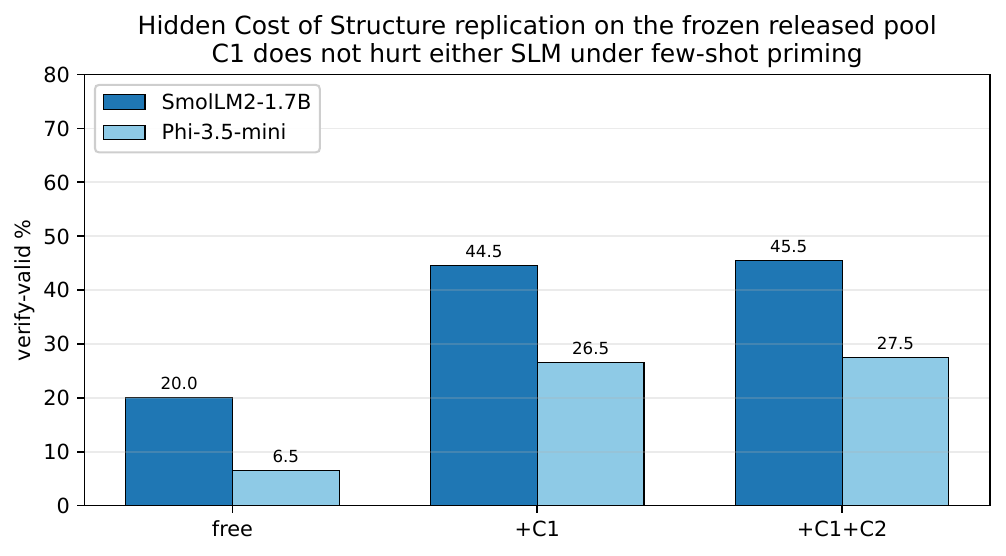}
\caption{Hidden Cost of Structure replication on the frozen arith+func
pool ($n{=}200$), with null result. $C_1$ is monotonically beneficial on
both SLMs ($+24.5$pp SmolLM2, $+20.0$pp Phi-3.5-mini, both $p<0.0001$).
Under 3-shot priming with a type-domain-aware grammar and bounded
identifiers, the RANLP 2025 HCS reversal does not reproduce.}
\label{fig:hcs}
\end{figure}

\begin{figure}[h]
\centering
\includegraphics[width=0.75\linewidth]{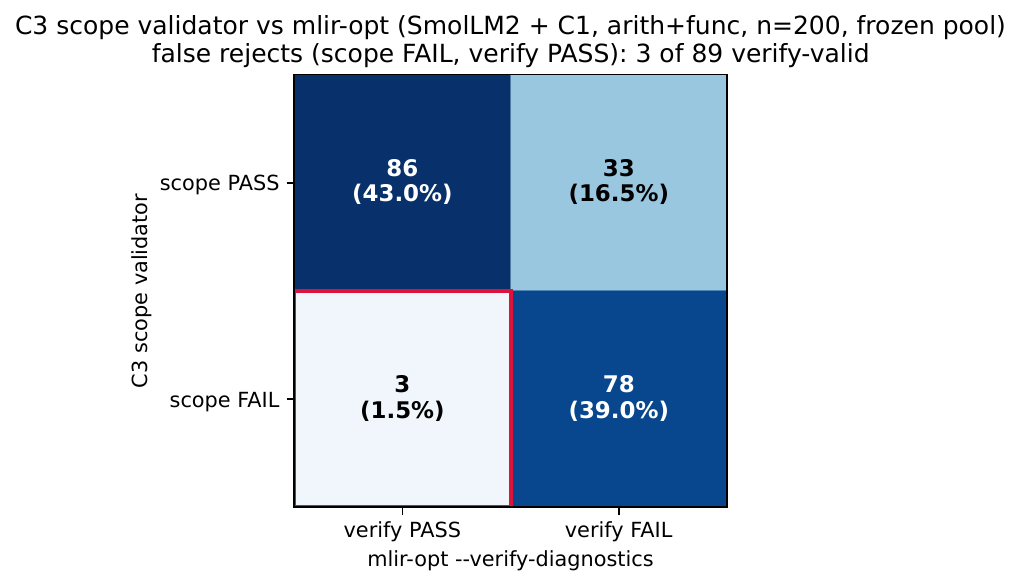}
\caption{$C_3$ scope-validator confusion matrix vs \texttt{mlir-opt} on
200 SmolLM2 + $C_1$ arith+func generations (frozen released pool). The
boxed (scope FAIL, verify PASS) cell counts false rejects: 3 of the 89
verify-valid programs. The same count is 0 for the $C_1$ outputs of every
15B/34B baseline on both dialects (Table~\ref{tab:c3_false_rejects}).}
\label{fig:scope}
\end{figure}

\begin{figure}[h]
\centering
\includegraphics[width=0.95\linewidth]{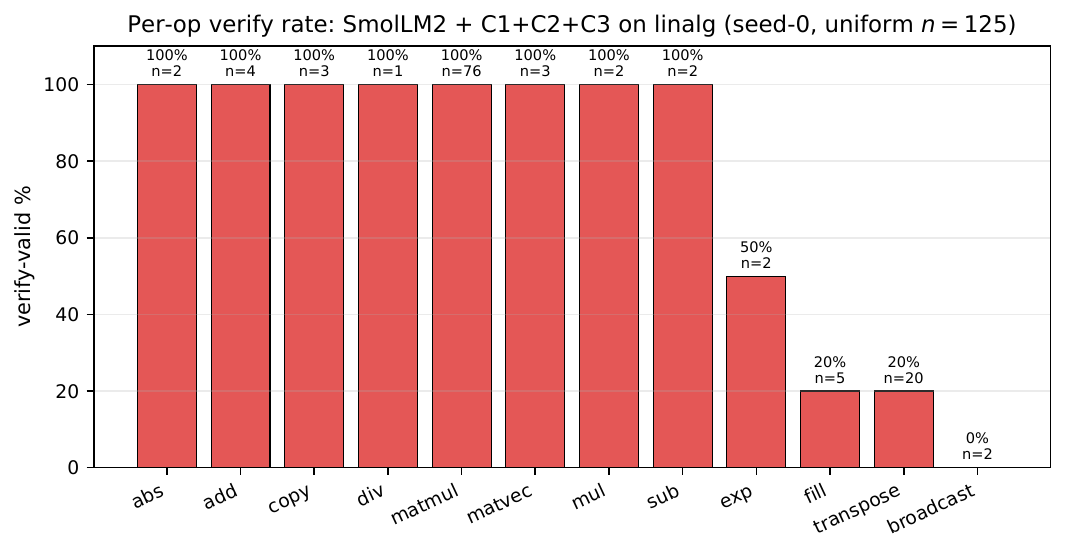}
\caption{Per-op verify rate under SmolLM2 + $C_1$+$C_2$+$C_3$ on
linalg (seed-0 run of the three-run frozen-pool cell, $n{=}125$). Most named
ops reach 100\%; shape-reasoning ops (\texttt{transpose},
\texttt{broadcast}) and the small \texttt{fill} and \texttt{exp} cells
trail, the residual gap motivating a future $C_4$ shape-inference
layer.}
\label{fig:perop}
\end{figure}

This appendix provides extended quantitative results referenced from
the main text: the headline cross-dialect comparison
(Fig.~\ref{fig:main}), the efficiency frontier
(Fig.~\ref{fig:efficiency}), the seed-0 ladder run with 95\% bootstrap
CIs (Table~\ref{tab:main}), the $C_3$ confusion matrix
(Fig.~\ref{fig:scope}), best-configuration baselines
(Table~\ref{tab:bestconfig}), both aggregations
(Table~\ref{tab:unique}), apples-to-apples paired comparisons
(Fig.~\ref{fig:apples}), per-dialect error-category progression
(Fig.~\ref{fig:err}, Table~\ref{tab:err}), paired-bootstrap deltas
(Fig.~\ref{fig:paired}), the HCS replication (Fig.~\ref{fig:hcs}),
the per-op residual-error breakdown (Fig.~\ref{fig:perop}), three-run
uniform-$n$ verify rates (Table~\ref{tab:multiseed}), the
empirical-claims summary (Table~\ref{tab:empirical_claims}), and a
note on run-to-run reproduction.

\begin{table}[h]
\centering
\caption{Constraint-layer progression on the frozen released pools
(seed-0 run; $n{=}200$ arith+func, $n{=}125$ linalg; verify-valid rate
with 95\% bootstrap CI in brackets). SmolLM2 rows form the constraint
ladder; baseline rows use five-retry $C_1$ rejection without $C_3$, and
``free'' is a single unconstrained attempt. \emph{Reading guide:}
compare adjacent SmolLM2 rows for each layer's contribution (paired
deltas in \S\ref{sec:ablations}); cross-system claims use the three-run
means in Table~\ref{tab:apples}.}
\label{tab:main}
\begin{tabular}{l r r}
\toprule
System & arith+func (n=200) & linalg (n=125) \\
\midrule
SmolLM2-1.7B, free & 20.0 [14.5, 25.5] & 52.0 [43.2, 60.8] \\
SmolLM2-1.7B + C1 & 44.5 [37.5, 51.5] & 64.0 [55.2, 72.0] \\
SmolLM2-1.7B + C1+C2 & 45.5 [38.5, 52.5] & 73.6 [65.6, 80.8] \\
\textbf{SmolLM2-1.7B + C1+C2+C3 (ours)} & 52.0 [45.0, 59.0] & 77.6 [70.4, 84.8] \\
Granite-Code-34B, free & 39.0 [32.5, 46.0] & 27.2 [20.0, 35.2] \\
Granite-Code-34B + C1 & 47.0 [40.5, 54.0] & 33.6 [25.6, 41.6] \\
CodeLlama-34B + C1 & 60.5 [53.5, 67.0] & 56.8 [48.0, 65.6] \\
\bottomrule
\end{tabular}

\end{table}

\begin{table}[h]
\centering
\caption{Best-configuration baselines on the frozen released pools:
verify-valid rate without $C_3$ (five-retry $C_1$ rejection, seed-0 run)
and with $C_3$ (mean of three runs). $C_3$ never lowers a baseline beyond
noise, so no baseline is represented by a $C_3$-penalized score; against
each baseline's best configuration SmolLM2's linalg lead (80.0\%) is at
least $+21.3$pp.}
\label{tab:bestconfig}
\begin{tabular}{l r r r r}
\toprule
System & arith C1-only & arith C1+C3 & linalg C1-only & linalg C1+C3 \\
\midrule
CodeLlama-34B & 60.5 & 59.8 & 56.8 & 58.7 \\
StarCoder2-15B & 67.0 & 66.8 & 47.2 & 54.9 \\
Granite-Code-34B & 47.0 & 51.5 & 33.6 & 35.7 \\
\bottomrule
\end{tabular}

\end{table}

\begin{table}[h]
\centering
\caption{Both aggregations for every pooled cell (verify-valid \%). ``All''
counts every prompt row; ``unique'' gives each unique prompt string one
value (the mean over its duplicates), per run, then averages runs. The
pools hold 193 (arith+func) and 75 (linalg) unique strings; StableHLO
pools contain no duplicates, so both aggregations coincide there. The
linalg ranking is unchanged (smallest SmolLM2 margin $+12.1$pp) and
arith+func remains a non-win.}
\label{tab:unique}
\small
\begin{tabular}{l r r r r}
\toprule
Cell & arith+func all & arith+func unique & linalg all & linalg unique \\
\midrule
SmolLM2 free & 20.0 & 18.7 & 52.0 & 38.7 \\
SmolLM2 + C1 & 44.5 & 43.5 & 64.0 & 56.0 \\
SmolLM2 + C1+C2 & 45.5 & 44.6 & 73.6 & 69.3 \\
SmolLM2 + C1+C2+C3 (seed-0 ladder) & 52.0 & 50.8 & 77.6 & 75.0 \\
\midrule
SmolLM2 + C1+C2+C3 (3 runs) & 53.2 & 52.2 & 80.0 & 75.7 \\
CodeLlama-34B + C1+C3 & 59.8 & 59.4 & 58.7 & 63.6 \\
StarCoder2-15B + C1+C3 & 66.8 & 67.2 & 54.9 & 40.4 \\
Granite-Code-34B + C1+C3 & 51.5 & 53.0 & 35.7 & 38.6 \\
Granite-Code-8B-fp16 + C1+C3 & 68.2 & 67.4 & 50.1 & 43.6 \\
\midrule
CodeLlama-34B + C1 & 60.5 & 60.1 & 56.8 & 61.3 \\
StarCoder2-15B + C1 & 67.0 & 67.4 & 47.2 & 38.7 \\
Granite-Code-34B + C1 & 47.0 & 48.7 & 33.6 & 37.0 \\
Granite-Code-34B free & 39.0 & 40.4 & 27.2 & 30.3 \\
\bottomrule
\end{tabular}

\end{table}

\begin{table}[h]
\centering
\caption{Error categorization across constraint levels, SmolLM2-1.7B
seed-0 ladder on the frozen released pools. Arity and dialect-misuse
buckets are 0 in every cell and are omitted.}
\label{tab:err}
\begin{tabular}{l l r r r r r}
\toprule
Dialect & Constraint & passed & type & type\_ssa & syntax & other \\
\midrule
arith+func & none & 40 & 1 & 3 & 0 & 156 \\
arith+func & c1 & 89 & 13 & 51 & 0 & 47 \\
arith+func & c1\_c2 & 91 & 1 & 61 & 0 & 47 \\
arith+func & c1\_c2\_c3 & 104 & 9 & 37 & 0 & 50 \\
\midrule
linalg & none & 65 & 0 & 1 & 4 & 55 \\
linalg & c1 & 80 & 0 & 5 & 19 & 21 \\
linalg & c1\_c2 & 92 & 0 & 7 & 0 & 26 \\
linalg & c1\_c2\_c3 & 97 & 0 & 4 & 0 & 24 \\
\bottomrule
\end{tabular}

\end{table}

\begin{table}[h]
\centering
\caption{Per-seed verify-valid rates across the apples-to-apples
cells. Four systems $\times$ two dialects, plus a
Granite-Code-8B-fp16 same-family precision control on both dialects,
at $n{=}200$ (arith+func) or $n{=}125$ (linalg) per run, three runs
(seeds 0, 1, 2). Baseline decoding is fully seeded; SmolLM2's first
attempt is greedy and its temperature-0.8 retries are not seeded, so its
three runs are repeats with measured variance. Mean is the across-run
average; $\pm$half-range is $(\max - \min)/2$. SmolLM2's seed-0
arith+func cell equals Table~\ref{tab:main}'s top rung (52.0\%); on
linalg the ladder run's 77.6\% lies within the measured run-to-run
spread (at most 2.4pp).}
\label{tab:multiseed}
\small
\begin{tabularx}{\linewidth}{@{}Lrrrrr@{}}
\toprule
Cell                                                     & s0      & s1       & s2       & mean    & $\pm$hr      \\
\midrule
SmolLM2 + $C_1$+$C_2$+$C_3$, arith+func                  & 52.0\%  & 52.0\%   & 55.5\%   & 53.2\%  & $\pm 1.8$pp  \\
SmolLM2 + $C_1$+$C_2$+$C_3$, linalg                      & 80.0\%  & 80.0\%   & 80.0\%   & 80.0\%  & $\pm 0.0$pp  \\
Granite-Code-34B-Q4 + $C_1$+$C_3$, arith+func            & 49.5\%  & 54.5\%   & 50.5\%   & 51.5\%  & $\pm 2.5$pp  \\
Granite-Code-34B-Q4 + $C_1$+$C_3$, linalg                & 34.4\%  & 40.8\%   & 32.0\%   & 35.7\%  & $\pm 4.4$pp  \\
CodeLlama-34B-Q4 + $C_1$+$C_3$, arith+func               & 60.5\%  & 59.0\%   & 60.0\%   & 59.8\%  & $\pm 0.8$pp  \\
CodeLlama-34B-Q4 + $C_1$+$C_3$, linalg                   & 59.2\%  & 60.8\%   & 56.0\%   & 58.7\%  & $\pm 2.4$pp  \\
StarCoder2-15B-Q4 + $C_1$+$C_3$, arith+func              & 68.0\%  & 65.0\%   & 67.5\%   & 66.8\%  & $\pm 1.5$pp  \\  
StarCoder2-15B-Q4 + $C_1$+$C_3$, linalg                  & 53.6\%  & 55.2\%   & 56.0\%   & 54.9\%  & $\pm 1.2$pp  \\
\midrule
Granite-Code-8B-fp16 + $C_1$+$C_3$, arith+func (qtz.\ control) & 67.5\%  & 71.5\%   & 65.5\%   & 68.2\%  & $\pm 3.0$pp  \\  
Granite-Code-8B-fp16 + $C_1$+$C_3$, linalg (qtz.\ control)     & 49.6\%  & 48.8\%   & 52.0\%   & 50.1\%  & $\pm 1.6$pp  \\
\bottomrule
\end{tabularx}
\end{table}

\begin{table}[h]
\centering
\caption{Summary of empirical claims and supporting evidence.}
\label{tab:empirical_claims}
\small
\begin{tabularx}{\linewidth}{@{}>{\raggedright\arraybackslash}p{0.25\linewidth}>{\raggedright\arraybackslash}p{0.17\linewidth}L@{}}
\toprule
Claim                                       & Status              & Supporting evidence \\
\midrule
linalg win (system-level)                   & Robust              & Mean of three runs 80.0\% (each run 80.0\%); $+21.3$ to $+44.3$pp over the baselines with non-overlapping CIs; at least $+21.3$pp over each baseline's best configuration; the fp16 control trails by 29.9pp \\
Enforcement asymmetry                       & Measured            & SmolLM2 under the baselines' post-hoc protocol: 24.5\% vs 52.0\% with masking on arith+func ($+27.5$pp, CI $[+20.0, +35.0]$) \\
8--25$\times$ per-generation speedup        & Robust              & Wall-clock on a single Apple M4 Max (Fig.~\ref{fig:efficiency}); 1.65--1.86 s vs 16--40 s \\
Mechanical schema port across dialects      & Robust              & Zero new constraint-layer code from arith+func+memref+linalg $\to$ StableHLO \\
StableHLO win on hand-curated Spec-30       & Corpus-conditional  & $+26.6$ to $+30.0$pp over the 34B baselines on Spec-30 only; baselines saturate at 98--100\% on Held-Out-200 \\
arith+func win                              & Not supported       & 53.2\% vs CodeLlama-34B 59.8\% (within CI), StarCoder2-15B 66.8\% (non-overlapping), Granite-Code-8B-fp16 68.2\% (paired $+15.0$pp) \\
Functional correctness                      & Partial             & Gold-differential on the spec prompts: 63.2\% (43/68) arith+func, 36.4\% (8/22) linalg under the canonical interface; adapted wrappers on the $n{=}10$ linalg references: 7 of 8 \\
\bottomrule
\end{tabularx}

\end{table}

\paragraph{Qualitative example (Linalg-Spec-30 prompt 14, ``transpose
a 3x5xf32 memref'').} SmolLM2 + $C_1$+$C_2$+$C_3$ emits:
\begin{lstlisting}[basicstyle=\small\ttfamily,frame=single,breaklines=true]
module { func.func @t(%A : memref<?x?xf32>, %B : memref<?x?xf32>) {
    linalg.transpose ins(%A : memref<?x?xf32>)
      outs(%B : memref<?x?xf32>) permutation = [1, 0]
    return
} }
\end{lstlisting}
which passes \texttt{mlir-opt --verify}. CodeLlama-34B + $C_1$ instead
invents a hand-rolled \texttt{linalg.generic} with malformed indexing
maps (\texttt{(d0, d1)[s0 -> s0 * 5 + d1]}) that fails parse. This
illustrates $C_2$'s role: the grammar restricts \texttt{linalg.*} to
the 12 named ops in scope, so SmolLM2 cannot emit
\texttt{linalg.generic} even by few-shot imitation.

\paragraph{Run-to-run reproduction.} Every table sits on the same
frozen pools. Re-running SmolLM2's full stack reproduces every prompt
resolved by the greedy first attempt byte-identically (109/109 on
arith+func, 105/105 on linalg); variation is confined to prompts that
reach the temperature-0.8 retries and spans at most 2.4pp across four
linalg runs (80.0, 80.0, 80.0, 77.6). Baseline runs are seeded: under
the pinned Ollama runtime, re-generating 20 published CodeLlama outputs
reproduces all 20 byte-identically.

\clearpage
\section{Datasheet for Datasets}
\label{sec:datasheet}

We follow the Gebru et al.~\cite{gebru2021datasheets}
Datasheet-for-Datasets template. The full datasheet ships in the
repository (\texttt{docs/datasheets/datasheet.md}); a condensed
version follows.

\paragraph{Motivation.} Natural-language-to-MLIR (``NL$\to$MLIR'')
is an under-represented task in existing code-LM benchmarks.
Existing MLIR corpora (e.g., the LLVM in-tree \texttt{mlir/test/})
are designed for compiler regression testing, not NL pairing;
HumanEval/MBPP target general-purpose code. Our four benchmarks fill
this gap. Their composition, together with the out-of-grammar stress
set and the functional references, is summarized in
Table~\ref{tab:per_benchmark_composition}.

\begin{table}[h]
\centering
\small
\caption{Per-benchmark composition. Spec-* benchmarks were
author-written against public ODS specs and verifier-filtered;
Held-Out-200 was generated by a parametric sweep and verifier-filtered;
OutOfGrammar-25 was hand-authored against StableHLO
ops the grammar does not cover; MLIR-Functional-Reference-30 adds
hand-authored execution references to 30 Spec prompts and is used to bound the verify-valid $\to$
output-match gap.}
\label{tab:per_benchmark_composition}
\small
\setlength{\tabcolsep}{4pt}
\begin{tabularx}{\linewidth}{@{}lrLlL@{}}
\toprule
Benchmark                   & $n$  & Dialect                                            & Source            & Type                                                \\
\midrule
MLIR-Spec-150               & 150  & arith+\allowbreak func+\allowbreak memref                                  & author-curated    & in-grammar                                          \\
Linalg-Spec-30              &  30  & linalg (12 ops)                                    & author-curated    & in-grammar                                          \\
StableHLO-Spec-30           &  30  & stablehlo (10 fam.)                                & author-curated    & in-grammar                                          \\
StableHLO-Held-Out-200      & 200  & stablehlo                                          & parametric sweep  & in-grammar (held-out)                               \\
StableHLO-OutOfGrammar-25 &  25  & stablehlo                                          & author-curated    & out-of-grammar (stress)                             \\
MLIR-Functional-Reference-30     &  30  & arith+\allowbreak func+\allowbreak memref / linalg+\allowbreak memref / stablehlo (10 ea.) & author-curated & evaluation artifact (NL+\allowbreak MLIR+\allowbreak inputs+\allowbreak expected output) \\
\bottomrule
\end{tabularx}

\end{table}

\paragraph{Composition.}
\textit{MLIR-Spec-150}: 150 NL$\to$MLIR pairs,
arith+func+memref, difficulty mix 38/44/18\%
easy/medium/hard. All reference MLIR verify-clean under
\texttt{mlir-opt --verify-diagnostics}.
\textit{Linalg-Spec-30}: 30 pairs covering the 12 linalg named ops in
scope under memref semantics.
\textit{StableHLO-Spec-30}: 30 author-curated pairs covering 10 op
families with tensor semantics.
\textit{StableHLO-Held-Out-200}: 200 programs constructed by a
parametric sweep over 7 op families $\times$ 6 dtypes $\times$
multiple shape ranks, filtered to iree-compile-clean (585 candidates
$\to$ 200 kept). \textit{MLIR-Functional-Reference-30}: 30 references
(10 per dialect, each attached to an existing MLIR-Spec-150,
Linalg-Spec-30, or StableHLO-Spec-30 prompt) with canonical signature, concrete inputs, and expected
output. One StableHLO reference (\texttt{divide}, f64) declares
\texttt{8xf64} inputs while its gold source declares
\texttt{tensor<32xf64>}, and IREE demotes f64 to f32 by default; it is
excluded from gold-gated scoring and the record is left unchanged.

\paragraph{Leakage and memorization.} The 210 spec prompts (150
arith+func, 30 linalg, 30 StableHLO) and their reference modules were
hand-authored for this benchmark in April 2026, after the pretraining
cutoff of every evaluated model (2023--2024 releases), and were not
scraped from any corpus. No SmolLM2 generation reproduces its gold
module even after whitespace normalization (0 of 540 spec-prompt
generations across runs); the 44 matches among 2,160 generation-gold
pairs are baseline outputs of short canonical programs of one or two
operations. The padding prompts used only in the evaluation pools are
machine-generated descriptions of LLVM/MLIR test files; to the extent
such files appear in pretraining data, they are at least as likely to
be in the large public-code crawls behind the 15B--34B baselines as in a
1.7B model's corpus, so memorization would inflate the baselines and
work against the paper's claim. Conclusions are unchanged on unique
prompts (Table~\ref{tab:unique}).

\paragraph{Collection + preprocessing.} Pairs are hand-authored
(Spec-30/150) or parametrically generated and filtered (Held-Out-200).
Each JSON contains \texttt{\{id, nl, mlir, difficulty, dialect,
notes\}}. No human-subject data; no PII.

\paragraph{Intended uses.} Evaluating NL$\to$MLIR generation
systems (constrained or unconstrained) under a verifier-based pass-rate
 metric. Functional-correctness claims need an execution harness;
the released gold-differential harness provides one for the 180 spec
prompts, and verify-valid alone measures structural validity only.

\paragraph{Distribution + license.} All four benchmarks, generated
outputs, grammars, and code released under \textbf{Apache License
2.0} (SPDX: Apache-2.0). Per-prompt generations from the baseline
models remain subject to their source model's terms (Llama 2
Community License for CodeLlama, BigCode OpenRAIL-M v1 for
StarCoder2). The Dockerfile pins LLVM, IREE, and
llama.cpp to their respective upstream licenses.

\paragraph{Maintenance.} Releases are version-tagged on the GitHub
repository, and corrections land in the next tag; the first release is
\texttt{v1.0-camera-ready}. Issues are
tracked on the GitHub repository and the Hugging Face dataset discussion
pages.

\paragraph{Croissant metadata.} Each benchmark ships a Croissant record
(in the repository under \texttt{eval/benchmarks/}, and with each
Hugging Face dataset) containing both core and Responsible-AI fields
(per MLCommons
Croissant~1.1~\cite{croissant2024} + RAI extensions, with provenance):
\texttt{cite-as}, \texttt{license}, \texttt{distribution},
\texttt{recordSet}, \texttt{annotations} with difficulty labels, and
the \texttt{rai:dataBiases}/\texttt{rai:dataLimitations} fields
describing the ``author-curated'' and ``parametric-sweep'' sampling
biases explicitly.

\paragraph{Reuse scenarios.} Beyond evaluating the constraint stack
proposed in this paper, we anticipate four uses for these benchmarks.
(i)~\emph{Evaluating new SLMs on MLIR generation.} A researcher with
a new small open-weight code model can run the released decoder
against MLIR-Spec-150 and Linalg-Spec-30 to obtain directly
comparable verify-valid rates; the evaluation-only intended use asks
users not to fine-tune on these pairs.
(ii)~\emph{Evaluating retrieval-augmented MLIR generation.} The
pairs are short and self-contained, making them suitable as targets
for retrieval-augmented decoding studies that condition on
dialect-specific reference programs. (iii)~\emph{Evaluating
fine-tuned MLIR models trained on external corpora.}
StableHLO-Held-Out-200's parametric construction provides resistance
to memorization through its templated-sweep generation procedure;
researchers releasing fine-tuned MLIR models can use it as a
held-out evaluation set, provided they did not train on it.
(iv)~\emph{Evaluating future constraint-decoding methods.} The
OutOfGrammar-25 stress set, the functional references, and the
gold-differential harness together let a future method demonstrate (a) graceful
out-of-grammar degradation and (b) closing of the verify-to-functional
 gap, both of which the present paper leaves open.

\section{Reproducibility-protocol Details}
\label{sec:reproducibility-appendix}

This appendix documents the full reproducibility protocol including code,
data, experimental setup, statistics, metric validation, and per-prompt
artifacts, that the NeurIPS Paper Checklist's reproducibility
questions point to for evidence.

\paragraph{Code release.} All code is released under Apache-2.0 in the
public repository, frozen at tag \texttt{v1.0-camera-ready}
(\url{https://github.com/plawanrath/slm-to-mlir-constrained-emitter}).
The lock file \texttt{scripts/env/requirements.lock.txt} pins every
host-side dependency at an exact version and is the single source of
truth. The
pinned LLVM container is built from \texttt{scripts/env/Dockerfile.llvm}
(\texttt{docker compose -f scripts/env/docker-compose.yml up -d}), and the
repository README maps every table and figure to the script that
regenerates it from the per-prompt files. Each rerun script checks its
environment first (byte-identical regeneration of published outputs,
identical verifier output on cached programs).

\paragraph{Data release.} All four primary benchmarks (MLIR-Spec-150,
Linalg-Spec-30, StableHLO-Spec-30, StableHLO-Held-Out-200) plus the
StableHLO-OutOfGrammar-25 stress set and the hand-authored $n{=}30$
functional reference set are released. The functional set annotates
30 existing Spec prompts (10 per dialect) and is released as evaluation
evidence, not as a training or fine-tuning set. Each benchmark ships with a per-instance JSON schema
at \texttt{eval/benchmarks/<name>/schema.json} and a Croissant 1.1
metadata record at \texttt{eval/benchmarks/<name>/croissant.json} with
MLCommons Responsible-AI extension fields (\texttt{rai:dataBiases},
\texttt{rai:dataLimitations}, provenance) populated. The Datasheet for
Datasets at \texttt{docs/datasheets/datasheet.md} covers every
Gebru-template field, including explicit ``uses the dataset does not
support'' (fine-tuning, functional claims from verify-valid alone); the full
datasheet is in \S\ref{sec:datasheet}. License is Apache-2.0 (SPDX) for
both code and data; the evaluation-only intended use (a request, not a
license term) is recorded in the
datasheet's \texttt{uses\_not\_supported} field and in the per-benchmark
Croissant \texttt{rai:dataLimitations} field.

\paragraph{Experimental protocol.} SmolLM2-1.7B-Instruct served via
MLX-LM 0.31.2 on Apple Silicon (M4 Max, 128 GB unified memory) at fp16.
Outlines 1.2.12 with the llguidance backend for C1 (CFG-guided decoding)
and C2 (type-domain splits); rejection-sampling wrapper for C3 (5-retry
budget, post-hoc \texttt{c3\_scope.validate} filter). Decoding
parameters: first try greedy at \texttt{temperature=0.0}, retries at
\texttt{temperature=0.8}, \texttt{max\_tokens=600}. Every system receives
the same instruction and the same three in-context examples; example
selection is fixed per benchmark and stored in the repository.
\emph{Serving template per cell:} SmolLM2 receives the prompt inside its
ChatML template (\texttt{<|im\_start|>}\ldots\texttt{<|im\_end|>});
Phi-3.5-mini (HCS replication) inside its \texttt{<|system|>},
\texttt{<|user|>}, \texttt{<|assistant|>} template; and each Ollama
baseline through its own native chat template, applied by the serving
stack. Baselines (CodeLlama-34B-Instruct and Granite-Code-34B-Instruct
at Q4\_K\_M, StarCoder2-15B-Instruct at Q4\_0) are served via Ollama
(0.32.1 reproduces the published generations byte-identically) at
temperature 0.2 on the first attempt and 0.3 on retries, seed
$s + 1000(k-1)$ for attempt $k$, and
\texttt{max\_tokens=256}.\footnote{Rerunning all three baselines at
\texttt{max\_tokens=600} (frozen pools, three runs, both dialects,
protocol otherwise identical) leaves every cell statistically unchanged:
three-run means differ from the 256-token cells by at most 0.2pp, and
paired per-prompt deltas by at most 0.3pp with every paired 95\% CI
spanning zero. Capping SmolLM2's free decode at 256 tokens truncates 16
of 200 outputs, every one already invalid at 600, and leaves its
verify rate unchanged.} Granite-Code-8B-Instruct is served via MLX at
fp16 as the precision control. Runs use seeds 0, 1, 2: baseline decoding
is fully seeded, while SmolLM2's temperature-0.8 retries are not, so its
three runs are repeats with measured variance
(Appendix~\ref{app:extras}). Bootstrap seed
\texttt{numpy.random.default\_rng(0)}. Verifier toolchain pinned: LLVM
\texttt{llvmorg-19.1.7} (\texttt{mlir-opt}), iree-compiler 20241104.1068;
both pinned in \texttt{scripts/env/requirements.lock.txt}.

\paragraph{Statistical reporting.} 95\% paired-bootstrap percentile CIs
on every cell (10,000 resamples), paired by \texttt{prompt\_id} so each
resample preserves per-prompt pairing across constraint conditions.
One-sided paired-bootstrap p-values on every directional claim.
Apples-to-apples cells use three runs (seeds $\{0, 1, 2\}$);
``mean'' reports the arithmetic mean across runs and ``$\pm$half-range''
reports $(\max - \min)/2$. Power-analysis statement: at expected
baseline pass-rate $p{=}0.55$, $n{=}200$ yields 95\% bootstrap CI
half-width below $\pm 8$ pp; at $p{=}0.5$, $n{=}125$ below $\pm 9$ pp;
at $p{=}0.5$, $n{=}30$ below $\pm 18$ pp.

\paragraph{Metric validation.} Verifier-tool substitution validated:
\texttt{iree-compile --compile-to=input} 50/50 concordant against
upstream \texttt{stablehlo-opt v1.4.0} on a stratified $n{=}50$ sample
across six op families. Build commands and per-instance log in
Appendix~\ref{app:stablehlo_toolchain}. Empty-input verifier gotcha disclosed and
guarded: \texttt{iree-compile --compile-to=input} accepts empty stdin
as a valid empty \texttt{module \{\}} and returns success, so our
wrapper requires a \texttt{func.func @} substring before invoking
\texttt{iree-compile}. We recommend any reuser apply the same guard.
Version drift is measured for the StableHLO path: every released
StableHLO module was re-verified under iree-base-compiler 3.11.0rc
(March 2026, about 16 months past the pin) through the same wrapper;
all 230 in-scope modules (30 Spec-30 and 200 Held-Out-200 golds) verify
identically, and the only divergence is 1 of the 25 out-of-grammar
stress programs (a 2-D convolution the newer toolchain rejects).

\paragraph{Per-prompt artifacts.} Every per-prompt generation is
released as \texttt{.jsonl} (one line per prompt $\times$ run, with
raw decoder output, retry log, verifier verdict). Bootstrap artifacts
released: \texttt{results/day52/layer\_delta\_bootstrap.json}
(paired-bootstrap CIs and one-sided p-values for the
\S\ref{sec:ablations} layer-deltas),
\texttt{results/day51\_seed0\_n200/apples\_to\_apples.json} (the
\S\ref{sec:results} apples-to-apples cells with their per-prompt
pairing), the frozen-pool ladders, matched-budget reruns, converse cell,
functional evaluations, and drift measurement (\texttt{results/day53}
to \texttt{results/day61}), and the provenance labels
(\texttt{results/day61/provenance.jsonl}).

\section*{NeurIPS Paper Checklist}
\label{sec:checklist}

\begin{enumerate}

\item {\bf Claims}
    \item[] Question: Do the main claims made in the abstract and introduction accurately reflect the paper's contributions and scope?
    \item[] Answer: \answerYes{}
    \item[] Justification: The abstract and \S\ref{sec:intro} frame the
    headline as dialect-specific and system-level: SmolLM2-1.7B with
    token-level schema-derived constraints matches or exceeds 15B/34B
    baselines running post-hoc rejection on linalg (robust across three
    runs) and on hand-curated StableHLO-Spec-30 (corpus-conditional; the
    templated parametric Held-Out-200 saturates the baselines), while on
    arith+func we report parity-to-loss. The enforcement asymmetry is
    measured (+27.5pp masking contribution), and verify-valid is scoped
    as structural (36.4\% of verify-valid linalg generations are
    functionally correct under the canonical interface). The scope is
    restated in \S\ref{sec:results} and \S\ref{sec:limits}.

\item {\bf Limitations}
    \item[] Question: Does the paper discuss the limitations of the work performed by the authors?
    \item[] Answer: \answerYes{}
    \item[] Justification: \S\ref{sec:limits} enumerates: 30B ``$C_1$''
    is rejection sampling rather than token-level masked decoding (the
    asymmetry is measured); the structural-vs-functional gap (63.2\%
    arith+func and 36.4\% linalg of verify-valid generations are
    functionally correct under the canonical interface); arith+func is a
    parity-to-loss cell; the StableHLO claim is corpus-conditional;
    single SLM family, single hardware platform, single LLVM version pin
    (drift measured for the StableHLO path only), with a
    Granite-Code-8B-fp16 control on both dialects addressing the
    precision-vs-quantization confound; grammar coverage limited to
    the released dialects; the in-line coupled decoder is currently
    a theoretical companion rather than an empirical contribution.

\item {\bf Theory assumptions and proofs}
    \item[] Question: For each theoretical result, does the paper provide the full set of assumptions and a complete (and correct) proof?
    \item[] Answer: \answerYes{}
    \item[] Justification: Appendix~\ref{app:soundness} gives
    Theorem~\ref{thm:sound} (soundness), Theorem~\ref{thm:cov} (BPE-refined coverage), and
    Theorem~\ref{thm:eq} (equivalence between in-line and rejection-sampled
    decoders on the accepted-string set). Each carries an explicit
    proof and a remark on the ``vocab-covers-ASCII'' assumption
    needed for Theorem~\ref{thm:cov}'s character-oracle construction.

\item {\bf Experimental result reproducibility}
    \item[] Question: Does the paper fully disclose all the information needed to reproduce the main experimental results of the paper to the extent that it affects the main claims and/or conclusions of the paper (regardless of whether the code and data are provided or not)?
    \item[] Answer: \answerYes{}
    \item[] Justification: \S\ref{sec:reproducibility} and
    Appendix~\ref{sec:reproducibility-appendix} document the decoder configuration (MLX-LM 0.31.2
    with Outlines 1.2.12 + llguidance for SmolLM2; Ollama for the
    15B/34B baselines; 5-retry rejection budget; first try greedy at
    \texttt{temperature=0.0}, retries at \texttt{temperature=0.8};
    \texttt{max\_tokens=600} for the SmolLM2 cell and
    \texttt{max\_tokens=256} for the 15B/34B baseline cells; 3-shot
    priming format), the verifier toolchain pins (LLVM
    \texttt{llvmorg-19.1.7}, \texttt{iree-compiler 20241104.1068}),
    and the bootstrap protocol (10{,}000 resamples, percentile 95\%
    CI, \texttt{numpy.random.default\_rng(0)}). The lock file
    \texttt{scripts/env/requirements.lock.txt} is the single source
    of truth for environment versions.

\item {\bf Open access to data and code}
    \item[] Question: Does the paper provide open access to the data and code, with sufficient instructions to faithfully reproduce the main experimental results, as described in supplemental material?
    \item[] Answer: \answerYes{}
    \item[] Justification: Code at
    \url{https://github.com/plawanrath/slm-to-mlir-constrained-emitter},
    frozen at tag \texttt{v1.0-camera-ready}, whose release asset
    \texttt{paper\_results.tar.gz} holds every per-prompt generation;
    benchmarks in the Hugging Face collection
    \url{https://huggingface.co/collections/plawanrath/cross-dialect-mlir-benchmarks-neurips-2026-e-and-d}
    (six dataset URLs in Appendix~\ref{sec:reproducibility}). Apache-2.0 license for
    both code and data. Every per-prompt generation is released, and
    the repository README maps each table and figure to the script that
    regenerates it from those files.

\item {\bf Experimental setting/details}
    \item[] Question: Does the paper specify all the training and test details (e.g., data splits, hyperparameters, how they were chosen, type of optimizer) necessary to understand the results?
    \item[] Answer: \answerYes{}
    \item[] Justification: \S\ref{sec:exp} (Experimental Setup) and
    \S\ref{sec:reproducibility} cover model choices
    (SmolLM2-1.7B-Instruct primary; CodeLlama-34B,
    Granite-Code-34B, StarCoder2-15B baselines under 4-bit
    quantization via Ollama), the serving template per cell, in-context
    example selection, evaluation protocol per dialect, and the
    5-retry rejection-sampling budget. No training is involved (the
    work is training-free); all configurations are inference-time.

\item {\bf Experiment statistical significance}
    \item[] Question: Does the paper report error bars suitably and correctly defined or other appropriate information about the statistical significance of the experiments?
    \item[] Answer: \answerYes{}
    \item[] Justification: All cell-level results are reported with
    paired bootstrap 95\% percentile CIs (10{,}000 resamples).
    Apples-to-apples comparisons (Table~\ref{tab:multiseed}) use three
    runs with mean $\pm$ across-run half-range. Per-layer deltas
    (\S\ref{sec:ablations}) are paired over identical prompts and
    reported with CIs that explicitly exclude or include zero; the
    unpaired minimum detectable effects are stated in \S\ref{sec:exp}.

\item {\bf Experiments compute resources}
    \item[] Question: For each experiment, does the paper provide sufficient information on the computer resources (type of compute workers, memory, time of execution) needed to reproduce the experiments?
    \item[] Answer: \answerYes{}
    \item[] Justification: \S\ref{sec:results} reports SmolLM2
    wall-clock as 1.65--1.86~s/generation; baselines
    16--40~s/generation. All experiments run on a single Apple M4
    Max laptop (no GPU cluster). The longest single experimental
    run (3-seed baseline linalg multi-seed, 750 samples
    across three 34B/15B models) took 5.4~hours wall-clock. Total
    compute budget for the paper is reproducible on a single
    64GB-RAM Apple Silicon machine, ${\sim}20$ days of wall-clock
    inference time across all systems; no cluster-scale training
    runs and no pre-training or fine-tuning.

\item {\bf Code of ethics}
    \item[] Question: Does the research conducted in the paper conform, in every respect, with the NeurIPS Code of Ethics \url{https://neurips.cc/public/EthicsGuidelines}?
    \item[] Answer: \answerYes{}
    \item[] Justification: No human subjects, no PII, no
    crowdsourced labor, no offensive content. The benchmark domain
    is technical MLIR code and English descriptions of numerical
    operations. All evaluated models are open-weight and accessed
    via their published licenses.

\item {\bf Broader impacts}
    \item[] Question: Does the paper discuss both potential positive societal impacts and negative societal impacts of the work performed?
    \item[] Answer: \answerYes{}
    \item[] Justification: \emph{Intended use and positive impact.}
    NL$\to$MLIR generation is a productivity tool for compiler
    engineers; the released benchmarks enable laptop-scale
    constrained code generation for compiler infrastructure and
    are intended for method evaluation, not as training data for
    safety-critical model deployment. \emph{Foreseeable risks.}
    (i) \emph{False-pass risk}: a verify-valid generation may
    still compute the wrong function; users should not treat
    verify-valid output as correctness-certified
    (\S\ref{sec:limits}, Table~\ref{tab:verify_vs_functional}).
    (ii) \emph{Benchmark contamination}: MLIR-Spec prompts are
    short NL descriptions of common ops; if memorized by a
    pretrained model, verify-valid rates would rise without
    reflecting generalization. (iii) \emph{Pretraining-provenance
    ambiguity}: there is no auditable line between
    ``MLIR seen during pretraining'' and verify-valid generation
    at inference; absolute rates should be read as artifacts of
    pretraining plus the constraint stack, not as model-intrinsic
    capability. \emph{Responsible release.} Benchmarks contain
    only public MLIR programs derived from public dialect specs;
    no PII and no human-subject data. We evaluate only a few
    open-weight models on a deliberate scope-of-evaluation basis,
    not as an ethical claim about excluded families; the method
    transfers to any open-weight model with an MLX or llama.cpp
    implementation.

\item {\bf Safeguards}
    \item[] Question: Does the paper describe safeguards that have been put in place for responsible release of data or models that have a high risk for misuse (e.g., pre-trained language models, image generators, or scraped datasets)?
    \item[] Answer: \answerNA{}
    \item[] Justification: The released artifacts are MLIR
    benchmarks and a constrained-decoding wrapper around publicly
    available open-weight models. No new pretrained model weights
    are released. We assess the misuse risk of structurally-valid
    IR generation as low.

\item {\bf Licenses for existing assets}
    \item[] Question: Are the creators or original owners of assets (e.g., code, data, models), used in the paper, properly credited and are the license and terms of use explicitly mentioned and properly respected?
    \item[] Answer: \answerYes{}
    \item[] Justification: All six evaluated models are credited
    and used, through Hugging Face Hub or Ollama and without
    redistributing weights, under their published terms:
    SmolLM2-1.7B-Instruct~\cite{allal2025smollm2} and the
    Granite-Code-8B fp16 control and Granite-Code-34B
    baseline~\cite{mishra2024granite} (Apache-2.0),
    Phi-3.5-mini-Instruct (MIT; \S\ref{sec:ablations} only),
    StarCoder2-15B~\cite{lozhkov2024starcoder2} (BigCode
    OpenRAIL-M v1), and CodeLlama-34B~\cite{roziere2023codellama}
    (Llama 2 Community License). The toolchain is MLIR/LLVM and
    IREE (Apache-2.0 with LLVM Exceptions), StableHLO and Outlines
    (Apache-2.0), and llguidance, MLX, Ollama, and llama.cpp (MIT),
    all compatible with our Apache-2.0 release.

\item {\bf New assets}
    \item[] Question: Are new assets introduced in the paper well documented and is the documentation provided alongside the assets?
    \item[] Answer: \answerYes{}
    \item[] Justification: A Gebru-style datasheet for the released
    benchmarks lives at \texttt{docs/datasheets/datasheet.md} and
    is summarized in \S\ref{sec:reproducibility} and the Datasheet
    section. Per-benchmark Croissant 1.1 metadata records (core and
    Responsible-AI fields, with provenance) ship at
    \texttt{eval/benchmarks/<name>/croissant.json} and with each
    Hugging Face dataset; the gold-differential harness ships with its
    README and per-prompt results. License: Apache-2.0 (SPDX:
    Apache-2.0).

\item {\bf Crowdsourcing and research with human subjects}
    \item[] Question: For crowdsourcing experiments and research with human subjects, does the paper include the full text of instructions given to participants and screenshots, if applicable, as well as details about compensation (if any)?
    \item[] Answer: \answerNA{}
    \item[] Justification: No crowdsourcing was used. All benchmark
    prompts and reference MLIR programs were authored or generated
    by the submitting author and verifier-filtered.

\item {\bf Institutional review board (IRB) approvals or equivalent for research with human subjects}
    \item[] Question: Does the paper describe potential risks incurred by study participants, whether such risks were disclosed to the subjects, and whether Institutional Review Board (IRB) approvals (or an equivalent approval/review based on the requirements of your country or institution) were obtained?
    \item[] Answer: \answerNA{}
    \item[] Justification: No human-subjects research.

\item {\bf Declaration of LLM usage}
    \item[] Question: Does the paper describe the usage of LLMs if it is an important, original, or non-standard component of the core methods in this research? Note that if the LLM is used only for writing, editing, or formatting purposes and does \emph{not} impact the core methodology, scientific rigor, or originality of the research, declaration is not required.
    \item[] Answer: \answerYes{}
    \item[] Justification: Language models are the subject of
    evaluation, not an unstated tool: the paper proposes a
    constrained-decoding scaffold around SmolLM2-1.7B-Instruct
    (primary), Phi-3.5-mini-Instruct (HCS replication only), and
    three baselines (CodeLlama-34B, Granite-Code-34B,
    StarCoder2-15B) under 4-bit quantization. All decoder
    configurations, prompting formats, and rejection-sampling
    budgets are specified in \S\ref{sec:exp} and Appendix~\ref{sec:reproducibility-appendix}. No
    LLM-generated text appears in the released benchmarks; all
    reference MLIR programs are author-curated and
    verifier-filtered. Ideation and research were done by the author.
    AI was used as a coding tool only. The paper was written by the
    author. After writing, AI was used for language tuning and minor
    edits.

\end{enumerate}

\end{document}